\documentclass[10pt]{article} % For LaTeX2e
\usepackage[preprint]{tmlr}
\usepackage{amsmath,amsfonts,bm}

\def\eqref#1{equation~\ref{#1}}
\def\1{\bm{1}}

\DeclareMathAlphabet{\mathsfit}{\encodingdefault}{\sfdefault}{m}{sl}
\SetMathAlphabet{\mathsfit}{bold}{\encodingdefault}{\sfdefault}{bx}{n}

\newcommand{\R}{\mathbb{R}}

\newcommand{\sigmoid}{\sigma}

\DeclareMathOperator*{\argmax}{arg\,max}
\DeclareMathOperator*{\argmin}{arg\,min}

\DeclareMathOperator{\sign}{sign}

\usepackage{hyperref}
\usepackage{url}

\usepackage[sets,operations,nn,colors,tikz]{cora-macs}
\usetikzlibrary{fadings,shadings} % For additive opacity in tikz.
\usepackage{pgfmath}
\pgfkeys{/pgf/declare function={arctanh(\x) = 0.5*(ln(1+\x)-ln(1-\x));}}

\usepackage{amsmath}
\usepackage{amsthm}
\usepackage{amssymb}
\usepackage{mathtools}
\usepackage{bbm} % for indicator function

\newtheorem{definition}{Definition}
\newtheorem{proposition}{Proposition}

\newtheorem*{rep@theorem}{\rep@title}
\newcommand{\newreptheorem}[2]{%
  \newenvironment{rep#1}[1]{%
    \def\rep@title{#2 \ref{##1}}%
    \begin{rep@theorem}}%
      {\end{rep@theorem}}}
\makeatother
\newreptheorem{proposition}{Proposition}
\newreptheorem{corollary}{Corollary}
\newreptheorem{theorem}{Theorem}

\let\originalleft\left
\let\originalright\right
\renewcommand{\left}{\mathopen{}\mathclose\bgroup\originalleft}
\renewcommand{\right}{\aftergroup\egroup\originalright}

\usepackage{nicefrac, xfrac}

\usepackage{aligned-overset}

\usepackage[inline]{enumitem}

\usepackage[linesnumbered,ruled,vlined]{algorithm2e}
\SetKw{Continue}{continue}

\usepackage{cleveref}
\crefname{figure}{Fig.}{Fig.}
\Crefname{figure}{Fig.}{Fig.}
\crefname{section}{Sec.}{Sec.}
\Crefname{section}{Sec.}{Sec.}
\crefname{equation}{}{}
\Crefname{equation}{}{}
\crefname{definition}{Def.}{Def.}
\Crefname{definition}{Def.}{Def.}
\crefname{proposition}{Prop.}{Prop.}
\Crefname{proposition}{Prop.}{Prop.}
\crefname{corollary}{Col.}{Col.}
\Crefname{corollary}{Col.}{Col.}
\crefname{theorem}{Thm.}{Thm.}
\Crefname{theorem}{Thm.}{Thm.}
\crefname{table}{Tab.}{Tab.}
\Crefname{table}{Tab.}{Tab.}
\crefname{algorithm}{Alg.}{Alg.}
\Crefname{algorithm}{Alg.}{Alg.}
\crefname{line}{line}{lines}
\Crefname{line}{line}{lines}
\crefrangeformat{line}{lines #3#1#4--#5#2#6}
\crefname{appendix}{Appendix}{Appendix}
\Crefname{appendix}{Appendix}{Appendix}
\crefformat{footnote}{#2\footnotemark[#1]#3}

\usepackage{booktabs}
\usepackage{makecell}
\usepackage{multirow}
\usepackage{colortbl}

\usepackage{wrapfig} 
\newcommand{\changed}[1]{{#1}}

\newcommand{\eye}[1]{\ensuremath{I_{#1}}} % identity matrix
\newcommand{\abs}[1]{\left|#1\right|}% absolute value
\newcommand{\zeros}{\ensuremath{\mathbf{0}}}
\newcommand{\ones}{\ensuremath{\mathbf{1}}}
\newcommand{\cmat}[1]{\begin{bmatrix} #1 \end{bmatrix}} % compat (small) matrix notation
\DeclareMathOperator\arctanh{\ensuremath{tanh^{-1}}}
\DeclareMathOperator{\area}{area}
\NewDocumentCommand{\nablaOperator}{e{_^}}{%
  \mathop{}\!% \mathop for good spacing before \nabla
  \nabla
  \IfValueT{#1}{_{\!#1}}% tuck in the subscript
  \IfValueT{#2}{^{#2}}% possible superscript
}
\newcommand{\grad}[2]{\nablaOperator_{#1}#2}

\newcommand{\partgrad}[2]{\frac{\partial #2}{\partial #1}}
\newcommand{\norm}[1]{\lVert#1\rVert}

\newcommand{\bigO}{\ensuremath{\mathcal{O}}}

\newcommand{\Weights}{\ensuremath{W}} % weight matrix
\newcommand{\bias}{\ensuremath{b}} % bias vector
\newcommand{\nnParams}{\ensuremath{\theta}}

\newcommand{\nnTargetLabel}{\ensuremath{l}} % target label

\newcommand{\approxFun}{\ensuremath{p}} % approximtion function
\newcommand{\xapproxError}{\ensuremath{x}}
\newcommand{\xapproxErrorL}{\ensuremath{\underline{\xapproxError}}}
\newcommand{\xapproxErrorU}{\ensuremath{\overline{\xapproxError}}}
\newcommand{\extremePoints}{\ensuremath{\mathcal{P}}}
\newcommand{\approxSlope}{\ensuremath{m}} % approximation slope
\newcommand{\approxOffset}{\ensuremath{d}} % approximation offset
\newcommand{\approxError}{\ensuremath{e}} % approximation error
\newcommand{\approxErrorL}{\ensuremath{\underline{\approxError}}} % lower approximation error
\newcommand{\approxErrorU}{\ensuremath{\overline{\approxError}}} % upper approximation error

\newcommand{\opBackpropEnclose}[3][]{\operatortt[#1]{backpropEnclose}{#2,#3}}

\newcommand{\singhSub}{\text{S}}

\DeclareMathOperator{\relu}{ReLU}

\title{Set-Based Training for Neural Network Verification}

\author{\name Lukas Koller \email lukas.koller@tum.de \\
      \addr Technical University of Munich \AND \\
      \name Tobias Ladner \email tobias.ladner@tum.de \\
      \addr Technical University of Munich \AND \\
      \name Matthias Althoff \email althoff@tum.de \\
      \addr Professorship for Cyber-Physical Systems \\
      Technical University of Munich}

\def\month{08}  % Insert correct month for camera-ready version
\def\year{2025} % Insert correct year for camera-ready version
\def\openreview{\url{https://openreview.net/forum?id=n0lzHrAWIA}} % Insert correct link to OpenReview for camera-ready version

\begin{document}

\maketitle

\begin{abstract}
  Neural networks are vulnerable to adversarial attacks, i.e., small input perturbations can significantly affect the outputs of a neural network.
  Therefore, to ensure safety of neural networks in safety-critical environments, the robustness of a neural network must be formally verified against input perturbations, e.g., from noisy sensors.
  To improve the robustness of neural networks and thus simplify the formal verification, we present a novel set-based training procedure in which we compute the set of possible outputs given the set of possible inputs and compute for the first time a gradient set, i.e., each possible output has a different gradient.
  Therefore, we can directly reduce the size of the output enclosure by choosing gradients toward its center.
  Small output enclosures increase the robustness of a neural network and, at the same time, simplify its formal verification.
  The latter benefit is due to the fact that a larger size of propagated sets increases the conservatism of most verification methods.
  Our extensive evaluation demonstrates that set-based training produces robust neural networks with competitive performance, which can be verified using fast (polynomial-time) verification algorithms due to the reduced output set.
\end{abstract}

% !TeX root = ../main.tex
% Add the above to each tex file to make compiling the PDF easier in some editors.

\section{Introduction} \label{sec:intro}

Neural networks demonstrate impressive performance for complex tasks, such as speech recognition~\citep{hinton_et_al_2012} or object detection~\citep{wang_et_al_2023}. However, many neural networks are sensitive to input perturbations~\citep{szegedy_et_al_2014}: Small, carefully chosen input perturbations can lead to unexpected outputs. This behavior is problematic for the adoption of neural networks in safety-critical environments, where the input often contains noisy sensor data or is subject to external disturbances, e.g., autonomous vehicle control~\citep{cunliang_et_al_2022} or airborne collision avoidance~\citep{irfan_et_al_2020}.
Thus, the formal verification of neural networks gained interest in recent years~\citep{brix_et_al_2023,brix_et_al_2024}. Given a set of inputs, the goal of formal verification of neural networks is to find a mathematical proof that the neural network returns the correct output for every input from the set. \changed{We focus on training and verifying robust neural networks. For the robustness verification, the goal is to verify that within bounded input perturbations, which are typically modeled as $\ell_\infty$-balls of radius $\epsilon\in\R_{>0}$, there is no adversarial input that is classified incorrectly.}

In this work, we address the challenge of training and verifying the robustness of a neural network with a novel set-based training procedure. \changed{Adversarially trained neural networks~\citep{madry_et_al_2018,zhang_et_al_2019} achieve great empirical robustness (against adversarial attacks). However, they remain hard to formally verify due to large approximation errors, which increase the size of enclosures of their output sets. Thus, we propose a novel set-based training that can explicitly enforce small output sets to simplify a subsequent formal verification. During training, we enclose the output set of the neural network and compute a gradient set, i.e., each possible output has a different gradient.
  We can directly enforce smaller output enclosures by choosing gradients that point toward the center of the enclosure.
  % Small output enclosures improve the robustness of the neural network and simplify subsequent formal verification.
}
To compute the gradient set, we use a set-based loss function, which considers the position (for accuracy) and the size of the output enclosure (for robustness).
Previous works are limited to a single gradient for training, thereby discarding much set-based information. \changed{\cref{fig:train-appr-comparison} compares our gradient set with the gradient of three robust training approaches: Interval bound propagation (IBP)~\citep{gowal_et_al_2019}, DiffAi~\citep{mirman_et_al_2018} and SABR~\citep{mueller_et_al_2023}. IBP computes bounds of the output set and uses the gradient of the worst-case output; DiffAi uses a zonotope enclosure to compute tighter bounds compared to IBP; SABR uses the gradient of the worst case of a smaller input region to reduce regularization by large approximation errors.}
Subsequently, we provide an overview of related work.

% \tikzexternaldisable % disable tikz external because of citation in figure.
\begin{figure}
  \centering
  \includetikz{gradient-comp/gradient-comparison}
  \caption{\changed{Comparing our gradient set with the gradient of other robust training approaches for the same neural network in the output space. The blue area shows the exact output set of the neural network. Other training approaches ((a) IBP~\citep{gowal_et_al_2019}, (b) DiffAi~\citep{mirman_et_al_2018} and (c) SABR~\citep{mueller_et_al_2023}) propagate intervals to compute a single gradient. On the contrary, our set-based training (d) computes a gradient set based on the position (for accuracy) and size (for robustness) of the output set.}}\label{fig:train-appr-comparison}
\end{figure}
% \tikzexternalenable % re-nable tikz-externalize}

\paragraph{Formal Verification of Neural Networks}
The formal verification of neural networks is computationally challenging, i.e., the problem is $\textsc{NP}$-hard with only rectified linear unit (ReLU) activation functions~\citep{katz_et_al_2017}; thus, even verifying small neural networks can take a long time.
Most formal verification approaches either
\begin{enumerate*}[label=(\roman*)]
  \item formulate an optimization problem, which is solved with an off-the-shelf solver, e.g., (mixed-integer) linear programming~\citep{singh_et_al_2019_boosting,zhang_et_al_2022,mueller_et_al_2022} or satisfiability modulo theories~\citep{katz_et_al_2017,wu_et_al_2024}, or
  \item use reachability analysis with efficient set representations, e.g., zonotopes~\citep{girard_2005}, and set-based computations to enclose the output set of a neural network (in polynomial time)~\citep{gehr_et_al_2018,singh_et_al_2019_abstract,wang_et_al_2021,kochdumper_et_al_2023,ladner_althoff_2023,lemesle_et_al_2024}. If the enclosure of the output set is sufficiently tight, it can be used to formally verify a neural network.
\end{enumerate*}
The challenge for most reachability-based approaches is the enclosure of nonlinearities in a neural network, which limits their scalability because each nonlinearity adds an approximation error that accumulates for large neural networks.
Therefore, in practice, often branch-and-bound procedures~\citep{brunel_et_al_2020,wang_et_al_2021,durand_et_al_2022,ferrari_et_al_2022} are used to recursively split the verification problem into smaller and simpler subproblems, e.g., by exploiting the piecewise linearity of the $\relu$ activation function. Due to the recursive process, branch-and-bound algorithms have exponential runtime in the worst case. We want to train robust neural networks that can be quickly verified without splitting the verification problem.

On the contrary, neural networks can be falsified by adversarial attacks, i.e., small input perturbations that lead to an incorrect output.
Often, adversarial attacks are fast to compute and effective at provoking incorrect outputs~\citep{goodfellow_et_al_2015}. The fast gradient sign method (FGSM) and projected gradient descent (PGD) are the most prominent approaches. The FGSM is a single-step gradient-based adversarial attack that efficiently generates adversarial attacks~\citep{goodfellow_et_al_2015}. PGD uses multiple iterations of FGSM to compute stronger adversarial attacks~\citep{kurakin_et_al_2017}. However, neural networks can be robust for one type of attack but can remain vulnerable to another type of attack~\citep{madry_et_al_2018}; thus, it is necessary to formally verify the robustness of neural networks.

\paragraph{Training Robust Neural Networks}
The training objective of a robust neural network is typically formulated as a min-max optimization problem~\citep{madry_et_al_2018}: minimize the worst-case loss within a set of possible inputs.
Computing the worst-case loss within a set is computationally difficult~\citep{weng_et_al_2018}. Nonetheless, robust neural networks can be effectively trained by approximating the worst-case loss with adversarial attacks, e.g., computed using PGD \citep{madry_et_al_2018}.

The well-established tradeoff-loss~\citep{zhang_et_al_2019} combines a regular loss for accuracy with a boundary loss for robustness, which is approximated using adversarial attacks. The boundary loss pushes the decision boundary of a classifier away from the training samples and thereby improves the robustness of the trained neural network. However, neural networks trained with adversarial attacks remain hard to formally verify.

Therefore, some approaches combine the training and formal verification of neural networks. In these works, the approximation of a worst-case loss is replaced by an upper bound, guaranteeing that no perturbation will lead to an incorrect output. Different methods for computing an upper bound of the worst-case loss have been proposed: Interval bound propagation (IBP)~\citep{gowal_et_al_2019}, linear relaxation~\citep{zhang_et_al_2020}, (mixed-integer) linear programming~\citep{wong_kolter_2018}, or abstract interpretation~\citep{mirman_et_al_2018}.
IBP computes conservative output bounds and uses the worst case within the bounds for training and verification~\citep{gowal_et_al_2019}~(\cref{fig:train-appr-comparison}a). However, large approximation errors can create an over-regularization and lead to poor performance~\citep{mueller_et_al_2023}.

Thus, by propagating smaller input sets, state-of-the-art robustness results are achieved~\citep{mueller_et_al_2023}~(\cref{fig:train-appr-comparison}b). However, a branch-and-bound algorithm with worst-case exponential-time complexity is used for their formal verification.
Instead, we aim to train robust neural networks that can be quickly verified using polynomial-time verification algorithms. Therefore, we enclose the output set of a neural network using zonotopes, which is significantly tighter than using IBP.
More closely related to our work is an approach using zonotopic enclosures during training~\citep{mirman_et_al_2018}; however, much set-based information is discarded by only using the enclosure to bound the worst-case loss. All related approaches for training robust neural networks consider a single gradient. Conversely, our approach considers a gradient set~(\cref{fig:train-appr-comparison}c), thereby simultaneously increasing the robustness of the neural network and simplifying the formal verification.

Other approaches combine IBP adversarial attacks computed in a latent space~\citep{mao_et_al_2023}. Similarly, zonotopes can be partially propagated through a neural network to compute adversarial attacks in latent space~\citep{balunovic_vechev_2020}; however, the partial propagation results in layer-wise training of the neural network, which has a significant computational overhead.
Furthermore, the training of robust neural networks can be improved by using a specialized initialization for the parameters of a neural network~\citep{shi_et_al_2021}.

\paragraph{Contributions}
Our main contributions are:
\begin{itemize}
  \item A novel set-based training procedure for robust neural networks that, for the first time, uses a gradient set for training that generalizes the well-established tradeoff-loss~\citep{zhang_et_al_2019}. The gradient set enables direct control of the size of the output enclosure to improve the robustness of the neural network and simplify formal verification.
  \item A fast, batch-wise, and differentiable set-based forward propagation and backpropagation that is efficiently computed on a GPU. The set propagation uses analytical solutions for the image enclosure of typical nonlinear activation functions.
  \item An extensive empirical evaluation in which we demonstrate the competitive performance of our set-based training and compare it with state-of-the-art robust training approaches. Moreover, we include large-scale ablation studies to justify our design choices.
\end{itemize}

\paragraph{Organization}
We introduce the required preliminaries in~\cref{sec:prelims}. In~\cref{sec:set-training}, we present our set-based training procedure, which benefits from a fast, batch-wise, and differentiable set propagation derived in~\cref{sec:fast_img_enc}. We provide an empirical evaluation, including ablation studies in~\cref{sec:evaluation}. Finally, we conclude our findings in~\cref{sec:concl}.

% !TeX root = ../paper.tex
% Add the above to each tex file to make compiling the PDF easier in some editors.

\section{Preliminaries}\label{sec:prelims}

\subsection{Notation}

Lowercase letters denote vectors and uppercase letters denote matrices.
The $i$-th entry of a vector $x$ is denoted by $x_{(i)}$.
For a matrix $A\in\R^{n\times m}$, $A_{(i,j)}$ denotes the entry in the $i$-th row and the $j$-th column, $A_{(i,\cdot)}$ denotes the $i$-th row, and $A_{(\cdot,j)}$ the $j$-th column.
The identity matrix is written as $\eye{n}\in\R^{n\times n}$. We use \zeros{} and \ones{} to represent the vector or matrix (with appropriate size) that contains only zeros or ones. Given two matrices $A\in\R^{m\times n_1}$ and $B\in\R^{m\times n_2}$, their (horizontal) concatenation is denoted by $\cmat{A\;B}\in\R^{m\times (n_1 + n_2)}$; if $n_1 = n_2$, their Hadamard product is the element-wise multiplication $(A\odot B)_{(i,j)} = A_{(i,j)}\,B_{(i,j)}$. The operation $\Diag{x}\in\R^{n\times n}$ returns a diagonal matrix with the entries of the vector $x\in\R^{n}$ on its diagonal.
We denote sets with uppercase calligraphic letters.
For a set $\mathcal{S}\subset\R^n$, we denote its projection to the $i$-th dimension by $\mathcal{S}_{(i)}$.
Given two sets $\mathcal{S}_1\subset\R^n$ and $\mathcal{S}_2\subset\R^m$, we denote the Cartesian product by $\mathcal{S}_1\times\mathcal{S}_2=\{\cmat{ s_1^\top\; s_2^\top}^\top\mid s_1\in\mathcal{S}_1,\,s_2\in\mathcal{S}_2\}$, and if $n = m$, we write the Minkowski sum as $\mathcal{S}_1\oplus\mathcal{S}_2=\{s_1 + s_2\mid s_1\in\mathcal{S}_1,\,s_2\in\mathcal{S}_2\}$.
For $n\in\N$, $[n]=\{1,2,\dots,n\}$ denotes the set of all natural numbers up to $n$.
An $n$-dimensional interval $\I\subset\R^n$ with bounds $l,u\in\R^n$ is denoted by $\I = \shortI{l}{u}$, where $\forall i\in[n]\colon l_{(i)}\leq u_{(i)}$.
For a function $f\colon\R^n\to\R^m$, we abbreviate its evaluation for a set $\mathcal{S}\subset\R^n$ with $f(\mathcal{S}) = \{f(s)\mid s\in\mathcal{S}\}$.
The derivative of a scalar function $f\colon\R\to\R$ is denoted as $f'(x)=\nicefrac{\mathrm{d}}{\mathrm{d} x}\,f(x)$.
Moreover, the gradient of a function $f\colon\R^n\to\R$ w.r.t. a vector $x\in\R^n$ is its element-wise derivative: $(\grad{x}{f(x)})_{(i)} = \nicefrac{\partial}{\partial x_{(i)}}\,\,f(x)$, for $i\in[n]$.
Analogously, we define the gradient of a function $f\colon\R^{n\times m}\to\R$ w.r.t. a matrix $A\in\R^{n\times m}$: $(\grad{A}{f(A)})_{(i,j)} = \nicefrac{\partial}{\partial A_{(i,j)}}\,\,f(A)$, for $i\in[n]$ and $j\in[m]$.

\subsection{Feed-Forward Neural Networks}\label{sec:prelim_nn_training}

A feed-forward neural network $\NN_\nnParams\colon\R^{\numNeurons_{0}}\to\R^{\numNeurons_\numLayers}$ consists of a sequence of $\numLayers\in\N$ layers. For the $k$-th layer, $\numNeurons_{k-1}\in\N$ denotes the number of input neurons and $\numNeurons_k\in\N$ denotes the number of output neurons. A layer can either be a linear layer, which applies an affine map, or a nonlinear (activation) layer, which applies a nonlinear activation function element-wise.
\begin{definition}[Neural Network, {\citep[Sec.~5.1]{bishop_2006}}]\label{def:nn_layers}
  For an input $\nnInput\in\R^{\numNeurons_0}$, the output $\nnOutput = \NN_\nnParams(\nnInput) \in\R^{\numNeurons_\numLayers}$ of a neural network $\NN_\nnParams$ is computed by
  \begin{align*}
    \nnHidden_0 & = \nnInput\text{,}                                                 &
    \nnHidden_k & = \nnLayer{k}{\nnHidden_{k-1}} \quad\text{for $k\in[\numLayers]$,} &
    \nnOutput   & = h_\numLayers\text{,}
  \end{align*}
  where
  \begin{equation*}
    \nnHidden_k = \nnLayer{k}{\nnHidden_{k-1}} = \begin{cases}
      \Weights_k\, \nnHidden_{k-1} + \bias_k & \text{if $k$-th layer is linear,} \\
      \actfun_k(\nnHidden_{k-1})             & \text{otherwise,}
    \end{cases}
  \end{equation*}
  with weights $\Weights_k\in\R^{\numNeurons_k\times \numNeurons_{k-1}}$, bias $\bias_k\in\R^{\numNeurons_k}$, and nonlinear activation function $\actfun_k$ which is applied element-wise.
\end{definition}
We denote the parameters of the neural network with $\nnParams$, which include all weight matrices and bias vectors from its linear layers.

\paragraph{Training of Neural Networks}
We consider supervised training settings of a classification task, where a neural network is trained on a dataset~$\mathcal{D}=\{(\nnInput_1,\nnTarget_1),\dots,(\nnInput_n,\nnTarget_n)\}$, containing inputs $\nnInput_i\in\R^{\numNeurons_0}$ with associated targets $\nnTarget_i\in\{0,1\}^{\numNeurons_\numLayers}$. A target $\nnTarget_i\in\{0,1\}^{\numNeurons_\numLayers}$ is a one-hot encoding of the target label $\nnTargetLabel_i\in[\numNeurons_\numLayers]$, i.e., $\nnTarget_{i(j)} = 1 \iff j = \nnTargetLabel_i$ for all $j\in[\numNeurons_\numLayers]$.
A loss function $\pointLoss:\R^{\numNeurons_\numLayers}\times\R^{\numNeurons_\numLayers}\to\R$ measures how well a neural network predicts the targets.
A typical loss function for classification tasks is the cross-entropy error:
\begin{align}
  \pointLoss_{\ceSub}(\nnTarget,\nnOutput) \coloneqq -\sum_{i=1}^{\numNeurons_\numLayers} \nnTarget_{(i)}\,\ln(p_{(i)})\text{,}
\end{align}
where $\ln$ denotes the natural logarithm and $p_{(i)}=\exp(\nnOutput_{(i)})/(\exp(\nnOutput)\,\ones)$ are the predicted class probabilities.
The training goal of a neural network is to find parameters $\nnParams$ that minimize the total loss of the dataset~$\mathcal{D}$~\citep[Sec.~5.2]{bishop_2006}:
\begin{equation}\label{eq:nn_training_goal}
  \min_\nnParams \sum_{(\nnInput_i,\nnTarget_i)\in\mathcal{D}} \pointLoss(\nnTarget_i,\NN_\nnParams(\nnInput_i))\text{.}
\end{equation}
We revisit the training of a neural network to later augment it with sets.
A popular algorithm to train a neural network is gradient descent~\citep[Sec.~5.2.4]{bishop_2006}: the parameters are randomly initialized~\citep{glorot_bengio_2010} and iteratively optimized using the gradient of the loss function.
We denote the gradient of the loss function $\pointLoss$ w.r.t. the output of the $k$-th layer $h_k$ as:
\begin{equation}\label{eq:backprop-grads}
  \nnGrad_{k} \coloneqq \grad{h_{k}}{\pointLoss(\nnTarget,\nnOutput)}\text{,}
\end{equation}
where $\nnOutput = \NN_\nnParams(\nnInput)$ for input $\nnInput\in\R^{\numNeurons_0}$. The weight matrix $\Weights_k$ and bias vector $\bias_k$ of the $k$-th layer are updated as \citep[Sec.~5.3]{bishop_2006}
\begin{align}\label{eq:point_weight_bias_update}
  \Weights_{k} & \gets \Weights_{k} - \eta\,\grad{\Weights_{k}}{\pointLoss(\nnTarget,\nnOutput)} = \Weights_{k} - \eta\,\nnGrad_{k}\,h^\top_{k-1}\text{,} &
  \bias_{k}    & \gets \bias_{k} - \eta\,\grad{\bias_{k}}{\pointLoss(\nnTarget,\nnOutput)} = \bias_{k} - \eta\,\nnGrad_{k} \text{,}
\end{align}
where $\eta\in\R_{>0}$ is the step size of gradient descent, i.e., the learning rate.
The gradients $g_k$ are efficiently computed with backpropagation~\citep[Sec.~5.3]{bishop_2006}: by utilizing the chain rule, the gradient $g_\numLayers$ of the last layer is propagated backward through all neural network layers.
\begin{proposition}[Backpropagation, {\citep[Sec.~5.3]{bishop_2006}}]\label{prop:point_backprop}
  Let $\nnOutput\in\R^{\numNeurons_\numLayers}$ be an output of a neural network with target $\nnTarget\in\R^{\numNeurons_\numLayers}$.
  The gradients $g_k$ are computed in reverse order as
  \begin{align*}
    \nnGrad_\numLayers & = \grad{\nnOutput}{\pointLoss(\nnTarget,\nnOutput)}\text{,}                                       &
    \nnGrad_{k-1}      & = \begin{cases}
                             \Weights_k^\top\, \nnGrad_k                             & \text{if $k$-th layer is linear,} \\
                             \Diag{\actfun'_k(h_{k-1}^{\phantom{\top}})}\, \nnGrad_k & \text{otherwise,}
                           \end{cases}
  \end{align*}
  for $k = \numLayers,\numLayers-1,\dots,1$.
\end{proposition}
From now on, we refer to this (standard) neural network training as point-based training.

\subsection{Set-Based Computation}

Our approach extends point-based training to sets, which we represent by zonotopes. A zonotope is a convex set representation describing the Minkowski sum of a finite number of line segments.
\begin{definition}[Zonotope, {\citep[Def.~1]{girard_2005}}]\label{def:zono}
  Given a center $c\in\R^n$ and a generator matrix $G\in\R^{n\times q}$, a zonotope $\Z\subset\R^n$ is defined as
  \begin{equation*}
    \Z = \left\{\left.c + \sum_{i=1}^{q} G_{(i,\cdot)}\,\beta_{(i)}\;\right|\;\beta\in [-\ones,\ones]\right\} \eqqcolon \shortZ{c}{G}\text{.}
  \end{equation*}
\end{definition}

Subsequently, we define several operations for zonotopes used in our training approach.
Please note that for the complexity analysis, we only consider the number of binary operations and neglect the computational effort of unary operations. Moreover, we consider the textbook method and do not assume any special numerical tricks that have been developed for large matrices.
\begin{proposition}[Interval Enclosure, {\citep[Prop.~2.2]{althoff_2010}}]\label{prop:interval_enclosure_zonotope}
  A zonotope $\Z = \shortZ{c}{G}$ with $c\in\R^n$ and $G\in\R^{n\times q}$ is enclosed by the interval $\shortI{l}{u}\supseteq\Z$, where
  \begin{align*}
    l & = c - \abs{G}\,\ones\text{,} & u & = c + \abs{G}\,\ones\text{,}
  \end{align*}
  and $\abs{\cdot}$ computes the element-wise absolute value. Computing an interval enclosure is in $\bigO(n\,q)$.
\end{proposition}
\begin{proposition}[Minkowski Sum, {\citep[Prop.~2.1 and Sec.~2.4]{althoff_2010}}]\label{prop:zono_mink_sum}
  The Minkowski sum of a zonotope $\Z = \shortZ{c}{G}$ and an interval $\I = \shortI{l}{u}\subset\R^n$ with $c,l,u\in\R^n$ and $G\in\R^{n\times q}$ is computed as
  \begin{equation*}
    \Z\oplus \I = \shortZ{c + \nicefrac{1}{2}\,(u + l)}{\cmat{G \; \nicefrac{1}{2}\,\Diag{u - l}}}\text{,}
  \end{equation*}
  and has time complexity $\bigO(n)$.
\end{proposition}
\begin{proposition}[Affine Map, {\citep[Sec.~2.4]{althoff_2010}}]\label{prop:linear_map_zonotope}
  The result of an affine map $f\colon\R^n\to\R^m,\,x\mapsto \Weights\,x + \bias$ with $\Weights\in\R^{m\times n}$ and $\bias\in\R^m$ applied to a zonotope $\Z = \shortZ{c}{G}$ with $c\in\R^n$ and $G\in\R^{n\times q}$ is
  \begin{equation*}
    f(\Z) = \left\{f(z)\mid z\in\Z\right\} = \Weights\,\Z \oplus \bias = \shortZ{\Weights\,c + b}{W\,G}\text{,}
  \end{equation*}
  and has time complexity $\bigO(m\,n\,q)$.
\end{proposition}
During set-based training, we want to reduce the size of the output sets. However, determining the volume of a zonotope is computationally demanding~\citep{elekes_1986}. Nevertheless, we can effectively approximate the size of a zonotope by its F-radius~\citep{combastel_2015}: The F-radius of a zonotope is the Frobenius norm of its generator matrix.
\begin{proposition}[F-Radius, {\citep[Def.~3]{combastel_2015}}]\label{prop:frad_norm}
  For a zonotope $\Z = \shortZ{c}{G} \subset\R^n$ with $G\in\R^{n\times q}$, the F-radius is
  \begin{equation*}
    \norm{\Z}_{F} \coloneqq \nicefrac{1}{n}\,\sqrt{\sum_{i=1}^{n}\sum_{j=1}^{q} G_{(i,j)}^2}\text{.}
  \end{equation*}
\end{proposition}
Please note that in contrast to \citep[Def.~3]{combastel_2015}, we include a normalization factor of $\nicefrac{1}{n}$.
For set-based training, we compute a gradient set, i.e., the gradient of a function w.r.t. a zonotope, which is also a zonotope, where the center is the derivative w.r.t. the center and the generator matrix is the derivative w.r.t. the generator matrix.
\begin{definition}[Zonotope Gradient]\label{def:zonotope_gradient}
  The gradient of a function $f(\cdot)$ w.r.t. a zonotope $\Z=\shortZ{c}{G}\subset\R^n$ is defined as
  \begin{equation*}
    \grad{\Z}{f(\Z)} \coloneqq \shortZ[]{\grad{c}{f(\Z)}}{\grad{G}{f(\Z)}}\text{.}
  \end{equation*}
\end{definition}

\begin{figure}
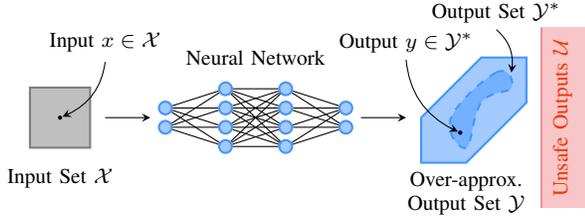

  \centering
  \includetikz{verify_nn}
  \caption{Verifying the local robustness of a neural network.}\label{fig:verify_nn}
\end{figure}

\subsection{Formal Verification of Neural Networks}\label{sec:verify_nn}

In this work, we consider the robustness of neural networks for classification tasks: Each dimension of an output $\nnOutput\in\R^{\numNeurons_\numLayers}$ corresponds to a class, and the dimension with the maximum value determines the predicted class.
An input $\nnInput\in\R^{\numNeurons_0}$ is correctly classified by a neural network if the predicted class matches the target label $\nnTargetLabel\in[\numNeurons_\numLayers]$:
\begin{equation}\label{eq:nn_robustness}
  \argmax_{k\in[\numNeurons_\numLayers]} \nnOutput_{(k)} = \nnTargetLabel\text{.}
\end{equation}
We call a neural network (locally) robust for a given set of inputs if the neural network correctly classifies every input within the set. Following previous works~\citep{madry_et_al_2018,gowal_et_al_2019,zhang_et_al_2019,mueller_et_al_2023}, we use the $\ell_\infty$-ball of radius $\nnPertRadius\in\R_{>0}$ around an input $\nnInput\in\R^{\numNeurons_0}$ as an input set:
\begin{equation}\label{def:eps_perturbed_inputs}
  \pi_\nnPertRadius(\nnInput) \coloneqq \shortZ{x}{\nnPertRadius\,\eye{\numNeurons_0}} = \left\{\left.\tilde{\nnInput}\in\R^{\numNeurons_0}\;\right|\;\norm{\tilde{\nnInput} - \nnInput}_\infty\leq\nnPertRadius\right\}\text{.}
\end{equation}

For a classification task with target label $\nnTargetLabel\in[\numNeurons_\numLayers]$, the unsafe set contains every incorrect classification, i.e., there is a dimension $k\in[\numNeurons_\numLayers]$ for which the output $\nnOutput_{(k)}$ is larger than the output of the target dimension $\nnOutput_{(l)}$~\citep[Prop.~B.2]{ladner_althoff_2023}:
\begin{equation}
  \nnUnsafeSet_\nnTargetLabel \coloneqq\left\{\nnOutput\in\R^{\numNeurons_\numLayers}\;\left|\;\exists k\in[\numNeurons_\numLayers]\colon \nnOutput_{(k)} > \nnOutput_{(l)}\right.\right\} = \changed{ \left\{\nnOutput\in\R^{\numNeurons_\numLayers}\;\left|\;(\eye{\numNeurons_\numLayers} - e_\nnTargetLabel^\top\,\ones)\,\nnOutput > 0\right.\right\}}\text{,}
\end{equation}
\changed{where $e_\nnTargetLabel\in\R^{\numNeurons_\numLayers}$ is the $l$-th standard basis vector.}
For an input set $\nnInputSet = \pi_\nnPertRadius(\nnInput)\subset\R^{\numNeurons_0}$, we formally verify the robustness of a neural network $\NN_\nnParams$ by using set-based computations to efficiently compute (in polynomial time) an enclosure $\nnOutputSet\subset\R^{\numNeurons_\numLayers}$ of its output set $\nnOutputSetExact\coloneqq \NN_\nnParams(\nnInputSet)\subseteq\nnOutputSet$~(\cref{fig:verify_nn}): If $\nnOutputSet$ does not contain unsafe outputs, we have formally verified the neural network for the input set $\nnInputSet$, as also $\nnOutputSetExact$ does not intersect with $\nnUnsafeSet$:
\changed{
  \begin{equation}
    \nnOutputSet\cap\nnUnsafeSet = \emptyset \implies \nnOutputSetExact\cap\nnUnsafeSet = \emptyset\text{.}
  \end{equation}
}
To compute an enclosure $\nnOutputSet$, we evaluate the layers (\cref{def:nn_layers}) over sets.
The output set of a linear layer is computed with an affine map~\citep[Sec.~2.4]{ladner_althoff_2023},
whereas the output set of a nonlinear layer is enclosed as it cannot be computed exactly for zonotopes, because they are not closed under nonlinear maps.
The required steps are summarized in~\cref{fig:img-enc-steps}. The activation function is applied element-wise; hence, the input dimensions are considered independently. We first project the input set onto a dimension and compute bounds (Steps~1 \& 2). The activation function is approximated using a linear function, and to ensure the soundness, a bound on the approximation errors is computed (Steps~3 \& 4). Finally, the approximation is evaluated over the input set, and the approximation errors are added (Steps~5 \& 6).

We define the following set-based forward propagation.
\begin{proposition}[Set-Based Forward Prop., {\citep[Sec.~2.4]{ladner_althoff_2023}}]\label{prop:set_forward_prop}
  For an input set $\nnInputSet\subset\R^{\numNeurons_0}$, an enclosure $\nnOutputSet\subset\R^{\numNeurons_\numLayers}$ of the output set $\nnOutputSetExact \coloneqq \NN_\nnParams(\nnInputSet)$ of a neural network can be computed as
  \begin{align*}
    \nnHiddenSet_0 & = \nnInputSet\text{,}                                                               &
    \nnHiddenSet_k & = \begin{cases}
                         \Weights_k\,\nnHiddenSet_{k-1} + \bias_k  & \text{if $k$-th layer is linear,} \\
                         \opEnclose{\actfun_k}{\nnHiddenSet_{k-1}} & \text{otherwise,}
                       \end{cases} \quad\text{for $k\in[\numLayers]$,} &
    \nnOutputSet   & = \nnHiddenSet_\numLayers \text{.}
  \end{align*}
  The operation $\opEnclose{\nnLayerName{k}}{\nnHiddenSet_{k-1}}$ encloses the image of a nonlinear layer~\citep[Prop.~2.14]{ladner_althoff_2023}.
  We denote the enclosure of the output set of a neural network $\NN_\nnParams$ by $\nnOutputSet = \opEnclose{\NN_\nnParams}{\nnInputSet}$.
\end{proposition}

\tikzexternaldisable % disable tikz external because of citation in figure.
\begin{figure}
  \centering
  \includetikz{img-enc-steps/main_steps_nonlinear}
  \caption{Main steps of an image enclosure~\citep[Prop.~2.14]{ladner_althoff_2023}.}\label{fig:img-enc-steps}
\end{figure}
\tikzexternalenable % re-nable tikz-externalize

\changed{
  Using the enclosure of the output set~(\cref{prop:set_forward_prop}), we can formally verify the robustness of a neural network.
  \begin{proposition}[Neural Network Verification, {\citep[Sec.~2.4]{ladner_althoff_2023}}]
    Given a neural network $\NN\colon\R^{\numNeurons_0}\to\R^{\numNeurons_\numLayers}$ and an input $\nnInput\in\R^{\numNeurons_0}$ with target label $\nnTargetLabel\in[\numNeurons_\numLayers]$, we enclose the output set $\nnOutputSet = \opEnclose{\NN_\nnParams}{\pi_\nnPertRadius(\nnInput)}$.
    We can formally verify the robustness of the neural network by 
    \begin{equation*}
      \max_{\nnOutput\in\nnOutputSet}\;(\eye{\numNeurons_\numLayers} - e_\nnTargetLabel^\top\,\ones)\,\nnOutput \leq 0 \implies \nnOutputSetExact\cap\nnUnsafeSet = \emptyset\text{.}
    \end{equation*}
  \end{proposition}
}

\subsection{Problem Statement}\label{sec:problem-statement}

\newcommand{\defSetTraining}{
  Given a set-based loss function $\setLoss:\R^{\numNeurons_\numLayers}\times 2^{\R^{\numNeurons_\numLayers}}\to \R$, set-based training trains a neural network using a gradient set $\grad{\nnOutputSet}{\setLoss(\nnTarget,\nnOutputSet)}\subset\R^{\numNeurons_\numLayers}$~(\cref{def:zonotope_gradient}), which defines a different gradient $\nnGrad\in\grad{\nnOutputSet}{\setLoss(\nnTarget,\nnOutputSet)}$ for each point $\nnOutput\in\nnOutputSet$.
}

We want to leverage set-based training to create robust neural networks.
\begin{definition}[Set-Based Training]\label{def:set-training}
  \defSetTraining
\end{definition}
Set-based training can directly enforce smaller output sets by choosing gradients that point toward the center of an output set~(\cref{fig:frad-grads}). Thereby, we can simultaneously improve the robustness of the neural network and simplify the formal verification.
We want to use set-based training to train the parameters $\nnParams$ of a robust neural network $\NN_\nnParams$:
\begin{equation*}
  \min_\nnParams \sum_{(\nnInput_i,\nnTarget_i)\in\mathcal{D}} \setLoss\bigl(\nnTarget_i,\opEnclose{\NN_\nnParams}{\pi_\nnPertRadius(\nnInput_i)}\bigr)\text{.}
\end{equation*}
By reducing the size of the output set using set-based training, we can verify neural networks with fast (polynomial time) algorithms.

% !TeX root = ../main.tex
% Add the above to each tex file to make compiling the PDF easier in some editors.

\section{Set-Based Training of Neural Networks}\label{sec:set-training}

We present a novel set-based algorithm to train robust neural networks with a gradient set~(\cref{def:set-training}).
In each training iteration, we
\begin{enumerate*}[label=(\roman*)]
  \item enclose the output set of the neural network by zonotopes~(\cref{subsec:set_forw_prop}),
  \item compute a gradient set derived from a set-based loss function using features of the output enclosure~(\cref{subsec:grad-set}),
  \item backpropagate the gradient set~(\cref{subsec:set_back_prop}), and
  \item aggregate the gradient set to update the parameters of the neural network~(\cref{subsec:set_weight_update}).
\end{enumerate*}

\subsection{Set-Based Loss Function and Gradient Set}\label{subsec:grad-set}

We derive the gradient set used for training by computing the gradient of a set-based loss function, which maps an output set to a loss value~(\cref{def:set-training}).
In this work, we define the set-based loss function $\setLoss:\R^{\numNeurons_\numLayers}\times 2^{\R^{\numNeurons_\numLayers}}\to \R$ so that it combines
\begin{enumerate*}[label=(\roman*)]
  \item a standard loss of the center of the enclosure (for accuracy) with
  \item the F-radius~(\cref{prop:frad_norm}) of the enclosure to approximate its size (for robustness)~(\cref{fig:train-appr-comparison}c).
\end{enumerate*}
\begin{definition}[Set-Based Loss]\label{def:set_loss_fun}
  We define a set-based loss function as
  \begin{align*}
    \setLoss(\nnTarget,\nnOutputSet) \coloneqq \underbrace{(1 - \tau)\,\pointLoss(\nnTarget,c_\numLayers)}_{\textup{Accuracy Loss}} + \underbrace{\nicefrac{\tau}{\nnPertRadius}\,\norm{\nnOutputSet}_F}_{\textup{Robustness Loss}}\text{,}
  \end{align*}
  where $\pointLoss\colon\R^{\numNeurons_\numLayers}\times\R^{\numNeurons_\numLayers}\to\R$ is a loss function, $\nnPertRadius\in\R_{>0}$ is the perturbation radius, and $\nnOutputSet = \shortZ{c_\numLayers}{G_\numLayers}$ is an output set.
\end{definition}
The set-based loss function balances the standard loss of the center and the F-radius of the output set using a hyperparameter $\tau\in\shortI{0}{1}$, controlling the tradeoff between accuracy and robustness. To make tuning the hyperparameter $\tau$ easier, the F-radius in~\cref{def:set_loss_fun} is normalized using the input perturbation radius $\nnPertRadius\in\R_{>0}$.
The normalization is derived by taking the ratio of the F-radii of output enclosure $\nnOutputSet$ and input set $\nnInputSet = \pi_\nnPertRadius(x)$, i.e., $\norm{\nnOutputSet}_F/\norm{\pi_\nnPertRadius(x)}_F \overset{\text{\cref{def:eps_perturbed_inputs}}}{=} \nicefrac{1}{\nnPertRadius}\,\norm{\nnOutputSet}_F$.

Our set-based loss generalizes the well-established tradeoff-loss \citep[Eq.~5]{zhang_et_al_2019}, which combines a standard training loss with a boundary loss; for $\nnOutput = \NN_\nnParams(\nnInput)$ and weighting factor $\lambda$:
\begin{equation}\label{eq:loss_trades}
  \pointLoss_{\text{TRADES}}(\nnTarget,\nnOutput) = \underbrace{\pointLoss(\nnTarget,\nnOutput)}_{\textup{standard training loss}} + \max_{\tilde{\nnInput}\in\pi_\nnPertRadius(\nnInput)} \underbrace{\nicefrac{1}{\lambda}\,\pointLoss(\nnOutput,\NN_\nnParams(\tilde{\nnInput}))}_{\textup{boundary loss}}\text{.}
\end{equation}

The F-radius in our set-based loss and the boundary loss in~\eqref{eq:loss_trades} have the same goal of training the robustness of the neural network. However, the F-radius captures the size of an output set in all directions, whereas the boundary loss only considers the distance to the furthest output.
Moreover, through the sound set-based computations (\cref{prop:set_forward_prop}), the set-based loss accurately over-approximates the size of the output set. In contrast, the boundary loss is only approximated in~\citep{zhang_et_al_2019} using adversarial attacks.

We derive the gradient set for training by computing the gradient of the set-based loss function~(\cref{def:set_loss_fun}), which requires the gradient of the F-radius.
\newcommand{\propGradFradNorm}{
  Given an output set $\nnOutputSet = \shortZ{c_\numLayers}{G_\numLayers}\subset\R^{\numNeurons_\numLayers}$, the gradient of the F-radius is
  \begin{equation*}
    \grad{\nnOutputSet}{\norm{\nnOutputSet}_{F}} = \shortZ[.]{\zeros}{\frac{1}{\numNeurons_\numLayers\,\norm{\nnOutputSet}_{F}}\,G_\numLayers}
  \end{equation*}
}
\begin{proposition}[Gradient of F-Radius]\label{prop:grad_frad_norm}
  \propGradFradNorm
  \begin{proof}
    See~\cref{app:sec:proofs}.
  \end{proof}
\end{proposition}
\begin{figure}
  \centering
  \includetikz{frad-grads}
  \caption{Gradients of the F-radius of a zonotope.}\label{fig:frad-grads}
\end{figure}
The negative gradients of the F-radius of a zonotope point toward the center of the zonotope (\cref{fig:frad-grads}); hence, minimizing the F-radius of a zonotope reduces the size of the zonotope.
With~\cref{prop:grad_frad_norm}, we compute the gradient set used for training:
\newcommand{\propGradSetLoss}{
  Given an output set $\nnOutputSet = \shortZ{c_\numLayers}{G_\numLayers}\subset\R^{\numNeurons_\numLayers}$, the gradient of the set-based loss function $\setLoss$ is
  \begin{equation*}
    \grad{\nnOutputSet}{\setLoss(\nnTarget,\nnOutputSet)} = \shortZ[.]{(1 - \tau)\,\grad{c_\numLayers}{\pointLoss(\nnTarget,c_\numLayers)}}{\frac{\tau}{\nnPertRadius}\,\frac{1}{\numNeurons_\numLayers\,\norm{\nnOutputSet}_{F}}\,G_\numLayers}
  \end{equation*}
}
\begin{proposition}[Set-Based Loss Gradient]\label{prop:grad_set_loss}
  \propGradSetLoss{}
  \begin{proof}
    See~\cref{app:sec:proofs}.
  \end{proof}
\end{proposition}
\cref{fig:train-appr-comparison}c visualizes the gradient set for samples from the output set. For effective set-based training using the gradient set, we require an efficient and differentiable set propagation, which we address next.

\section{Fast, Batch-wise, and Differentiable Set-Propagation}\label{sec:fast_img_enc}

We require efficient, batch-wise, and differentiable set propagation to implement set-based training. In this section, we first define a forward propagation with the required properties~(\cref{subsec:set_forw_prop}) before we derive the corresponding set-based backpropagation~(\cref{subsec:set_back_prop}) and weight updates~(\cref{subsec:set_weight_update}). Finally, we investigate the time complexity of our set-based training~(\cref{subsec:set_train_compl}).

\subsection{Set-Based Forward Propagation}\label{subsec:set_forw_prop}

The set propagation through linear layers is computed with an affine map, which can be efficiently implemented batch-wise using matrix multiplications on a GPU. Further, we want to efficiently compute batch-wise image enclosures of nonlinear layers.
Sampling-based methods~\citep{ladner_althoff_2023,kochdumper_et_al_2023} are impractical for this task because they use polynomial regression and thus are not efficient enough for backpropagation.
In contrast,~\citep{singh_et_al_2018} derives fast analytical solutions for the approximation errors of a specific linear approximation of s-shaped activation functions; however, this approach causes large approximation errors. During training, large approximation errors induce an over-regularization, leading to poor performance~\citep{mueller_et_al_2023}.
To address this issue, we derive analytical solutions for the approximation errors of linear approximations for three typical activation functions: rectified linear unit ($\relu$), hyperbolic tangent, and logistic sigmoid.
Secondly, we provide a linear approximation whose approximation errors are smaller or equal to Singh's enclosure~\citep[Thm.~3.2]{singh_et_al_2018} while being equally fast to compute.

For the remainder of this section, let $\actfun\colon\R\to\R$ be a monotonically increasing function, which is approximated by a linear function $\approxFun\colon\R\to\R,\,x\mapsto\approxSlope\,x + \approxOffset$ within an interval $\shortI{l}{u}\subset\R$.
The approximation errors of $\approxFun$ are given by the largest lower distance and upper distance between $\actfun$ and $\approxFun$ (\cref{fig:approx_error}).
\begin{definition}[Approximation Error of Linear Approximation]\label{def:approx_error} % , {\citep[Sec.~3]{singh_et_al_2018}}
  The approximation error of a linear approximation $\approxFun$ for $\actfun$ within the interval $\shortI{l}{u}$ are defined as
  \begin{align*}
    \approxErrorL & \coloneqq \min_{x\in\shortI{l}{u}} \actfun(x) - \approxFun(x)\text{,} & \approxErrorU & \coloneqq \max_{x\in\shortI{l}{u}} \actfun(x) - \approxFun(x)\text{.}
  \end{align*}
\end{definition}
We enclose the output of the activation function $\actfun$ as,
\begin{equation*}
  \actfun(\shortI{l}{u})\subseteq\approxFun(x) \oplus \shortI{\approxErrorL}{\approxErrorU}\text{.}
\end{equation*}
\cref{fig:approx_error} illustrates a linear approximation $\approxFun$ of the hyperbolic tangent along with its approximation errors.
% \tikzexternaldisable % disable tikz external because of citation in figure.
\begin{figure}
  \centering
  \includetikz{imgenc_group}
  \caption{Image enclosure of hyperbolic tangent: (left) Our linear approximation and approximation errors; (right) Comparison of our image enclosure and Singh's enclosure~\citep[Thm.~3.2]{singh_et_al_2018}.}\label{fig:approx_error}
\end{figure}
% \tikzexternalenable % re-nable tikz-externalize

\paragraph{Efficient Computation of Approximation Errors}\label{sec:approx_errors}

We efficiently find the approximation errors $\approxErrorL$ and $\approxErrorU$ of a linear approximation $\approxFun$ by only checking specific points of the interval $\shortI{l}{u}$, i.e., extreme points of the activation function $\actfun$.
\newcommand{\propApproxError}{
  The approximation error of $\approxFun$ for $\actfun\in\{\relu, \tanh, \sigmoid\}$ are computed as
  \begin{align*}
    \approxErrorL & = \min_{x\in\extremePoints_{\actfun}\cap\shortI{l}{u}} \actfun(x) - \approxFun(x)\text{,} &
    \approxErrorU & = \max_{x\in\extremePoints_{\actfun}\cap\shortI{l}{u}} \actfun(x) - \approxFun(x)\text{,}
  \end{align*}
  where
  \begin{align*}
    \extremePoints_{\relu}    & = \left\{0,l,u\right\}\text{,}                                                      &
    \extremePoints_{\tanh}    & = \left\{\pm\arctanh\left(\sqrt{1 - \approxSlope}\right),l,u\right\}\text{,}        &
    \extremePoints_{\sigmoid} & = \left\{\pm 2\,\arctanh\left(\sqrt{1 - 4\,\approxSlope}\right),l,u\right\}\text{.}
  \end{align*}
}
\begin{proposition}[Approximation Errors for ReLU, Hyperbolic Tangent, and Logistic Sigmoid]\label{prop:approx_error}
  \propApproxError
  \begin{proof}
    See~\cref{app:sec:proofs}.
  \end{proof}
\end{proposition}
\newcommand{\propReluApproxError}{
  The approximation errors of $\approxFun$ for ReLU are computed as
  \begin{align*}
    \approxErrorL & = \min_{x\in\extremePoints_{\relu}} \relu(x) - \approxFun(x)\text{,} \\
    \approxErrorU & = \max_{x\in\extremePoints_{\relu}} \relu(x) - \approxFun(x)\text{,}
  \end{align*}
  where $\extremePoints_{\relu} = \left\{l,0,u\right\}\cap\shortI{l}{u}$.
}
We note that our computation of the approximation errors works for any (monotonically increasing) linear approximation.
Moreover, we observe that the offset $\approxOffset$ of the linear approximation $\approxFun$ does not affect the image enclosure; hence, w.l.o.g. we set $\approxOffset = 0$.
\begin{definition}[Linear Approximation of an Activation Function]\label{def:lin_approx}
  Within the interval $\shortI{l}{u}$, we approximate an activation function $\actfun$ by a linear function $\approxFun(x) \coloneqq \approxSlope\, x$, where
  \begin{align*}
    \approxSlope & \coloneqq \frac{\actfun(u) - \actfun(l)}{u - l}\text{.}
  \end{align*}
\end{definition}
We note that the slope of our approximation is identical to the slope used for the triangle relaxation by DeepPoly~\citep{singh_et_al_2019_abstract}; however, instead of formulating linear constraints, we require the explicit computation~(\cref{prop:approx_error}) of the approximation errors for the zonotope propagation.
Our image enclosure no longer uses a polynomial regression or requires sampling to compute the approximation errors~\citep[Sec.~3.2]{kochdumper_et_al_2023}.
Moreover, only matrix operations are required, i.e., matrix multiplication, matrix addition, $\min$, and $\max$. Therefore, it can be efficiently computed batch-wise, taking full advantage of GPU acceleration.

\paragraph{Our Enclosure vs. Singh's Enclosure}

We motivate the choice for our image enclose by comparing it with Singh's enclosure~\citep[Thm.~3.2]{singh_et_al_2018}.
For s-shaped activation functions, e.g., hyperbolic tangent and logistic sigmoid, we prove that the approximation error of our linear approximation (\cref{def:lin_approx}) is always smaller or equal to the approximation error of Singh's enclosure~\citep[Thm.~3.2]{singh_et_al_2018} w.r.t. the area in the input-output plane (see~\cref{fig:approx_error}) measuring the integrated approximation error over $\shortI{l}{u}$:
\begin{equation}\label{eq:img_enc_area}
  \area(\shortI{\approxErrorL}{\approxErrorU},\shortI{l}{u}) \coloneqq (u - l)\,(\approxErrorU - \approxErrorL)\text{.}
\end{equation}
\newcommand{\thmImgEncAreaSmaller}{
  Let $\actfun$ be an s-shaped function, and let $\shortI{l}{u}$ be an interval. Moreover, let $\shortI{\approxErrorL}{\approxErrorU}$ be the approximation errors of our enclosure~(\cref{def:approx_error,def:lin_approx}), and let $\approxError_{\singhSub}$ be the approximation error of Singh's enclosure~\citep[Thm.~3.2]{singh_et_al_2018}.
  It holds that
  \begin{equation*}
    \area(\shortI{\approxErrorL}{\approxErrorU},\shortI{l}{u}) \leq \area(\shortI{-\approxError_{\singhSub}}{\approxError_{\singhSub}},\shortI{l}{u})\text{.}
  \end{equation*}
}
\begin{proposition}\label{thm:img_enc_area_smaller}
  \thmImgEncAreaSmaller
  \begin{proof}
    See~\cref{app:sec:proofs}.
  \end{proof}
\end{proposition}
\cref{fig:img_enc_output_sets} shows an instance where the output set computed with our image enclosure is significantly smaller and does not intersect the unsafe region. In contrast, the output set computed with Singh's enclosure intersects the unsafe region.
% \tikzexternaldisable
\begin{figure}
  \centering
  \includetikz{imgenc-our-vs-singh}
  \caption{Comparison of the output set of a neural network computed with our image enclosures and Singh's enclosure~\citep[Thm.~3.2]{singh_et_al_2018}.}\label{fig:img_enc_output_sets}
\end{figure}
% \tikzexternalenable

\subsection{Set-Based Backpropagation}\label{subsec:set_back_prop}

We now derive the corresponding set-based backpropagation.
Analogous to the standard backpropagation~(\cref{prop:point_backprop}), the set-based backpropagation computes for every layer of the neural network the gradient set w.r.t. the output set $\nnHiddenSet_k = \shortZ{c_{k}}{G_{k}}$:
\begin{equation}\label{eq:set-backprop-grads}
  \nnGradSet_k = \shortZ{c'_k}{G'_k} \coloneqq \grad{\nnHiddenSet_k}{\setLoss(\nnTarget,\nnOutputSet)}\text{.}
\end{equation}
The set-based backpropagation of linear layers is straightforward as it applies an affine map (\cref{prop:point_backprop}). However, the backpropagation of nonlinear layers is more involved: while the image enclosure only uses linear approximations, these depend on the input set. Thus, intuitively, we have to apply the product rule.

Please note that, we use the plus symbol ($+$) between zonotopes to denote their element-wise addition: $\shortZ{c_1}{G_1} + \shortZ{c_2}{G_2} = \shortZ{c_1 + c_2}{G_1 + G_2}$, whereas the Minkowski sum is denoted as $\shortZ{c_1}{G_1} \oplus \shortZ{c_2}{G_2} = \shortZ{c_1 + c_2}{\begin{bsmallmatrix} {\displaystyle G_1} & {\displaystyle G_2}\end{bsmallmatrix}}$.
\newcommand{\propBackpropFastEnclose}{
Assume the $k$-th layer is a nonlinear layer with activation function $\actfun_{k}$. Given an input set $\nnHiddenSet_{k-1} = \shortZ{c_{k-1}}{G_{k-1}}$ with $G_{k-1}\in\R^{\numNeurons_{k-1}\times p}$ and a gradient set $\nnGradSet_{k} = \shortZ{c'_{k}}{G'_{k}}$, the gradient set $\nnGradSet_{k-1}=\shortZ{c'_{k-1}}{G'_{k-1}}$ is computed for each dimension $i\in[\numNeurons_{k}]$ as
\begin{align*}
  \nnGradSet_{k-1(i)} & = \approxSlope_{k(i)}\,\nnGradSet_{k(i)} + \left(c'_{k(i)}\,c_{k-1(i)} + G'_{k(i,[p])}\,G_{k-1(i,\cdot)}^\top\right)\,\grad{\nnHiddenSet_{k-1(i)}}{\approxSlope_{k(i)}}                                                               \\
                      & \quad + \frac{1}{2}\,\left(c'_{k(i)} + G'_{k(i,p + i)}\right)\,\grad{\nnHiddenSet_{k-1(i)}}{\approxErrorU_{k(i)}} + \frac{1}{2}\,\left(c'_{k(i)} - G'_{k(i,p + i)}\right)\,\grad{\nnHiddenSet_{k-1(i)}}{\approxErrorL_{k(i)}}\text{.}
\end{align*}
The operation $\nnGradSet_{k-1}=\opBackpropEnclose{\actfun_{k}}{\nnGradSet_{k}}$ computes the gradient set of an image enclosure.
}
\begin{proposition}[Backpropagation through Image Enclosure]\label{prop:backprop_fast_enclose}
  \propBackpropFastEnclose
  \begin{proof}
    See~\cref{app:sec:proofs}.
  \end{proof}
\end{proposition}

Using the backpropagation of an image enclosure, we can (analogous to~\cref{prop:point_backprop}) backpropagate the gradient sets through all layers of a neural network.
\newcommand{\propSetBackprop}{
  The gradient sets $\nnGradSet_k$ are computed in reverse order as
  \begin{align*}
    \nnGradSet_\numLayers & = \grad{\nnOutputSet}{\setLoss(\nnTarget,\nnOutputSet)}\text{,}                     &
    \nnGradSet_{k-1}      & = \begin{cases}
                                W_k^\top\,                      \nnGradSet_k & \text{if $k$-th layer is linear,} \\
                                \opBackpropEnclose{\actfun_k}{\nnGradSet_k}  & \text{otherwise,}
                              \end{cases}
  \end{align*}
  for $k = \numLayers,\numLayers-1,\dots,1$.}
\begin{proposition}[Set-Based Backpropagation]\label{prop:set_backprop}
  \propSetBackprop
  \begin{proof}
    See~\cref{app:sec:proofs}.
  \end{proof}
\end{proposition}

\subsection{Set-Based Update of Weights and Biases}\label{subsec:set_weight_update}

We now describe how to use the gradient set $\nnGradSet_k$ and the set of inputs $\nnHiddenSet_{k-1}$ to update the weights and biases of a linear layer:
Intuitively, we compute the outer product between the gradient set $\nnGradSet_k$ and input set $\nnHiddenSet_{k-1}$.
To avoid clutter, we define the outer product between two zonotopes $\Z_1=\shortZ{c_1}{G_1}\subset\R^{n_1}$ and $\Z_2=\shortZ{c_2}{G_2}\subset\R^{n_2}$ as $\Z_1\odot\Z_2^\top\coloneqq c_1\,c_2^\top + G_1\,G_2^\top\in\R^{n_1\times n_2}$.
\newcommand{\propSetWeightUpdate}{
  The gradients of the set-based loss w.r.t. a weight matrix and a bias vector are
  \begin{align*}
    \grad{W_{k}}{\setLoss(\nnTarget,\nnOutputSet)} & = \grad{\nnHiddenSet_{k}}{\setLoss(\nnTarget,\nnOutputSet)}\odot\nnHiddenSet_{k-1}^\top\text{,} &
    \grad{b_{k}}{\setLoss(\nnTarget,\nnOutputSet)} & = \grad{\nnHiddenSet_{k}}{\setLoss(\nnTarget,\nnOutputSet)}\odot\shortZ{1}{\zeros}^\top\text{.}
  \end{align*}
}
\begin{proposition}[Gradient Set of Weights and Biases]\label{prop:set-weight-update}
  \propSetWeightUpdate
  \begin{proof}
    See~\cref{app:sec:proofs}.
  \end{proof}
\end{proposition}
The weight matrices and bias vectors are updated analogous to standard training~\cref{eq:point_weight_bias_update} using the gradients of the set-based loss function:
\begin{align}\label{eq:weight_bias_update_sets}
  W_{k} & \gets W_{k} - \eta\,\grad{W_{k}}{\setLoss(\nnTarget,\nnOutputSet)}\text{,} &
  b_{k} & \gets b_{k} - \eta\,\grad{b_{k}}{\setLoss(\nnTarget,\nnOutputSet)}\text{.}
\end{align}

\subsection{Computational Complexity}\label{subsec:set_train_compl}

Finally, we derive the time complexity of set-based training.
\cref{algo:fast_enclose} implements our image enclosure.
\begin{algorithm}
  \DontPrintSemicolon
  \SetInd{0.5em}{0.5em}
  \SetKwFunction{fastEncloseAlgo}{\fastEncloseName}
  \SetKwProg{fun}{function}{}{}

  \fun{$\opEnclose{\actfun_k}{\nnHiddenSet_{k-1}}$}{
  Find bounds $[l_{k-1},u_{k-1}]$ of $\nnHiddenSet_{k-1}$ \tcp*{\cref{prop:interval_enclosure_zonotope}}\label{line:alg_fast_img_enc:bounds}
  \For{$i \gets 1$ \KwTo $\numNeurons_k$}{
    Find linear approx. $\approxSlope_{k(i)}\,x$ of $\actfun_k$ \tcp*{\cref{def:lin_approx}} \label{line:alg_fast_img_enc:lin_approx}
    Find approx. errors $\approxErrorL_{k(i)}$, $\approxErrorU_{k(i)}$ \tcp*{\cref{prop:approx_error}}\label{line:alg_fast_img_enc:approx_errors}
  }
  $\widetilde{\nnHiddenSet}_{k} \gets \Diag{\approxSlope_k}\,\nnHiddenSet_{k-1} + \nicefrac{1}{2}\,(\approxErrorU_{k} + \approxErrorL_{k})$ \tcp*{\cref{prop:linear_map_zonotope}}\label{line:alg_fast_img_enc:lin_map}
  $\nnHiddenSet_{k} \gets \widetilde{\nnHiddenSet}_{k} \oplus \left[\approxErrorL_{k},\approxErrorU_{k}\right]$ \tcp*{\cref{prop:zono_mink_sum}}\label{line:alg_fast_img_enc:mink}
  \Return $\nnHiddenSet_k$\;
  }
  \caption{Image Enclosure of a Nonlinear Layer.}
  \label{algo:fast_enclose}
\end{algorithm}

We provide the time complexity of~\cref{algo:fast_enclose} w.r.t. the number of input dimensions and generators.
\newcommand{\propFastEncloseTimeComplexity}{
  For an input set $\nnHiddenSet_{k-1}=\Z$ with $c\in\R^n$ and $G\in\R^{n\times q}$,~\cref{algo:fast_enclose} has time complexity $\bigO(n^2\,q)$ w.r.t. the number of input dimensions $n$ and the number of generators $q$.
}
\begin{proposition}[Time Complexity of~\cref{algo:fast_enclose}]\label{prop:fast_enclose_time_complexity}
  \propFastEncloseTimeComplexity
  \begin{proof}
    See~\cref{app:sec:proofs}.
  \end{proof}
\end{proposition}
\newcommand{\propAlgoBackpropFastEnclose}{
  \cref{algo:backprop_fast_enclose} has time complexity $\bigO(\numNeurons_{k-1}\,\numNeurons_k\,q)$, w.r.t. the number of input neurons $\numNeurons_k$ and the number of generators $q$.
}
\newcommand{\propSetForwardPropComplexity}{
  The zonotopes during a forward propagation~(\cref{prop:set_forward_prop}) have at most $q \leq \numNeurons_0 + \sum_{k\in[\numLayers]}\numNeurons_k$ generators.
  Let $\numNeurons_\text{max} \coloneqq \max_{k\in[\numLayers]} \numNeurons_{k}$ be the maximum number of neurons in a layer.
  The set-based forward propagtion~(\cref{prop:set_forward_prop}) has time complexity $\bigO(\numNeurons_\text{max}^2\,q\,\numLayers)$ w.r.t. $\numNeurons_\text{max}$, $q$ and the number of layers $\numLayers$.
}
\changed{
  Using~\cref{prop:fast_enclose_time_complexity}, we can compute the time complexity of a set-based forward propagation.
  \begin{proposition}[Time Complexity of~\cref{prop:set_forward_prop}]\label{prop:set_forward_complexity}
    \propSetForwardPropComplexity
    \begin{proof}
      See~\cref{app:sec:proofs}.
    \end{proof}
  \end{proposition}}

\cref{algo:set_training} implements an iteration of set-based training.
\begin{algorithm}
  \DontPrintSemicolon
  \SetInd{0.5em}{0.5em}
  \KwData{Input $\nnInput\in\R^{\numNeurons_0}$, Target $\nnTarget\in\R^{\numNeurons_\numLayers}$}
  \KwResult{Neural Network with Updated Parameters}

  $\nnHiddenSet_0 \gets \shortZ{\nnInput}{\nnPertRadius\,\eye{\numNeurons_0}}$ \tcp*{construct input set \cref{def:eps_perturbed_inputs}}\label{line:alg_set_train:init_perturb_input}
  \For(\tcp*[f]{set-based forward prop. (\cref{prop:set_forward_prop})}){$k\gets1$ \KwTo $\numLayers$}{\label{line:alg_set_train:forward_prop_start}
    \eIf{$k$-th layer is linear}{
      $\nnHiddenSet_k \gets W_k\,\nnHiddenSet_{k-1} + b_k$\;\label{line:alg_set_train:forward_lin}
    }{
      $\nnHiddenSet_k \gets \opEnclose{\actfun_k}{\nnHiddenSet_{k-1}}$\;\label{line:alg_set_train:forward_act}\label{line:alg_set_train:forward_prop_end}
    }
  }
  $\nnOutputSet \gets \nnHiddenSet_\numLayers$\tcp*{obtain output set \cref{def:eps_perturbed_inputs}}\label{line:alg_set_train:output}
  $\nnGradSet_\numLayers \gets \grad{\nnOutputSet}{\setLoss(\nnTarget,\nnOutputSet)}$ \tcp*{compute gradient set (\cref{prop:grad_set_loss})}\label{line:alg_set_train:compute_loss}
  \For(\tcp*[f]{set-based backprop. (\cref{prop:set_backprop})}){$k \gets \numLayers$ \KwTo $1$}{\label{line:alg_set_train:back_prop_start}
    \eIf{$k$-th layer is linear}{
      $\nnGradSet_{k-1} \gets W_k^\top\,\nnGradSet_k$\;\label{line:alg_set_train:backward_lin}
      $W_k \gets W_k - \eta\,(\nnGradSet_k\odot\nnHiddenSet_{k-1}^\top)$\tcp*{\cref{prop:set-weight-update} and \eqref{eq:weight_bias_update_sets}}\label{line:alg_set_train:weight_bias_update_start}\label{line:alg_set_train:backward_weight_update}
      $b_k \gets b_k - \eta\,(\nnGradSet_k\odot\shortZ{1}{\zeros}^\top)$\tcp*{\cref{prop:set-weight-update} and \eqref{eq:weight_bias_update_sets}}\label{line:alg_set_train:backward_bias_update}\label{line:alg_set_train:weight_bias_update_end}
    }{
      $\nnGradSet_{k-1} \gets \opBackpropEnclose{\actfun_k}{\nnGradSet_k}$\tcp*{\cref{prop:backprop_fast_enclose}}\label{line:alg_set_train:backward_act}\label{line:alg_set_train:back_prop_end}
    }
  }
  \caption{Set-based Training Iteration. Hyperparameters: $\nnPertRadius\in\R_{>0}$, $\tau\in\shortI{0}{1}$, and $\eta\in\R_{>0}$.}
  \label{algo:set_training}
\end{algorithm}
First, we enclose the output set of the neural network (\crefrange{line:alg_set_train:init_perturb_input}{line:alg_set_train:forward_prop_end}). Then, we compute the gradient set (\cref{line:alg_set_train:compute_loss}), which is backpropagated through the neural network (\crefrange{line:alg_set_train:back_prop_start}{line:alg_set_train:back_prop_end}). Finally, the parameters of the neural network are updated (\crefrange{line:alg_set_train:weight_bias_update_start}{line:alg_set_train:weight_bias_update_end}).

\newcommand{\propSetTrainingComplexity}{
  % The zonotopes used in~\cref{algo:set_training} have at most $q \leq \numNeurons_0 + \sum_{k\in[\numLayers]}\numNeurons_k$ number of generators.
  % Let $\numNeurons_\text{max} \coloneqq \max_{k\in[\numLayers]} \numNeurons_{k-1}\,\numNeurons_k$ be the maximum size of a weight matrix in the neural network.
  % Moreover,~
  \cref{algo:set_training} has time complexity $\bigO(\numNeurons_\text{max}^2\,q\,\numLayers)$ w.r.t. $\numNeurons_\text{max}$, $q$ and the number of layers $\numLayers$.
}
\begin{proposition}[Time Complexity of~\cref{algo:set_training}]\label{prop:set_training_complexity}
  \propSetTrainingComplexity
  \begin{proof}
    See~\cref{app:sec:proofs}.
  \end{proof}
\end{proposition}

The time complexity of set-based training is polynomial, and compared to point-based training, only has an additional factor $q\in\bigO(\numNeurons_0 + \sum_{k\in[\numLayers]}\numNeurons_k)$. The increased time complexity is expected because set-based training propagates entire generator matrices through the neural network. Moreover, for some linear relaxation methods, similar time complexities are reported~\citep{zhang_et_al_2018}.

% !TeX root = ../paper.tex
% Add the above to each tex file to make compiling the PDF easier in some editors.
\section{Evaluation}\label{sec:evaluation}

We use the MATLAB toolbox CORA~\citep{althoff_2015} to implement set-based training.
Following previous works~\citep{gowal_et_al_2019}, we train a 6-layer convolutional neural network on \textsc{Mnist}~\citep{lecun_2010}, \textsc{Svhn}~\citep{netzer_et_al_2011}, \textsc{Cifar-10}~\citep{krizhevsky_2009}, and \textsc{TinyImageNet}~\citep{le_yang_2015}. The training details can be found in~\cref{sec:app:train-details}. We first present the main results and then justify our design choices with extensive ablation studies.

\subsection{Main Results}\label{sec:eval-main-results}

We compare our set-based training against adversarial tradeoff training, i.e., TRADES~\citep{zhang_et_al_2019}, as well as two state-of-the-art interval-based training approaches, i.e., IBP~\citep{gowal_et_al_2019} and SABR~\citep{mueller_et_al_2023}. For each training scheme, we report the clean accuracy (percentage of correctly classified test samples), the falsified accuracy (percentage of test samples for which PGD finds an adversarial example)~\citep{gowal_et_al_2019}, and, unlike previous works, the \changed{fast-verified accuracy (percentage of verified test samples using zonotopes). Most related works report verified accuracies computed with slow (exponential time) verification algorithms using branch-and-bound~\citep{balunovic_vechev_2020,mueller_et_al_2023,mao_et_al_2023} or mixed-integer programming~\citep{gowal_et_al_2019}; the verification can take up to 34h for \textsc{Mnist} networks~\citep[App. C]{mueller_et_al_2023}. Our goal is the fast (polynomial time) verification of neural networks. Thus, we report fast-verified accuracy, which uses a single zonotope propagation~(\cref{prop:set_forward_prop}) with polynomial time complexity~(\cref{prop:set_forward_complexity}) for the verification of each test input without a branch-and-bound procedure. The verification of a 6-layer convolutional neural network trained on \textsc{Mnist} with 10\,000 test inputs takes approximately 30min.}
\begin{table}[tp]
  % \footnotesize
  \centering
  \caption{Comparison with state of the art (mean \& std. dev. of the best $3$ runs across $5$ seeds).}\label{tab:main_results_table}
  \begin{tabular}{lcl >{$}r<{$}@{$\,\pm\,$}>{$}l<{$} >{$}r<{$}@{$\,\pm\,$}>{$}l<{$} >{$}r<{$}@{$\,\pm\,$}>{$}l<{$} >{($}l<{$)}}
    \toprule
    \bf{Dataset}                           & $\mathstrut \nnPertRadius_\infty$    & \bf{Method} & \multicolumn{2}{c}{\bf{clean Acc.}} & \multicolumn{2}{c}{\bf{falsified Acc.}} & \multicolumn{3}{c}{\bf{fast-verified Acc. (max)}}                                          \\ \midrule
    % [top-3 of 5] TRADES: ./tars/results/mnist-cnn-med-dec-09-1718-dec10-0104-merge/ (with grad clipping); IBP, SABR, Set: ./tars/results/mnist-cnn-med-dec05/,  (no grad clipping)
    \multirow{4}{*}{\textsc{Mnist}}        & \multirow{4}{*}{$0.1$}               & TRADES      & \bf{99.40}                          & 0.04                                    & \bf{98.40}                                        & 0.06  & 0.00       & 0.00 & 0.00       \\
                                           &                                      & IBP         & 97.76                               & 0.26                                    & 96.32                                             & 0.39  & 95.58      & 0.41 & 96.05      \\ % ./tars/results/mnist-ibp-sabr-set-1
                                           &                                      & SABR        & 97.94                               & 0.44                                    & 95.84                                             & 1.28  & 92.80      & 3.57 & 95.86      \\% ./tars/results/mnist-ibp-sabr-set-1
                                           &                                      & Set (ours)  & 98.76                               & 0.29                                    & 97.52                                             & 0.26  & \bf{95.89} & 0.82 & \bf{96.40} \\ % ./tars/results/mnist-ibp-sabr-set-1
    \midrule
    % Option: TRADES, IBP, SABR: ./results/241231-022334
    % [top-3 of 5] TRADES, IBP, SABR: ./tars/results/cifar10-cnn-med-ibp-sabr-trades-dec27-1918-1-5, Set: ./tars/results/cifar10-cnn-med-set-dec21-2228
    \multirow{4}{*}{\textsc{Cifar-10}}     & \multirow{4}{*}{$\nicefrac{2}{255}$} & TRADES      & \bf{82.96}                          & 0.35                                    & \bf{67.35}                                        & 0.01  & 0.00       & 0.00 & 0.00       \\
                                           &                                      & IBP         & 46.54                               & 2.92                                    & 42.13                                             & 2.56  & 36.83      & 1.74 & 38.80      \\ % ./results/cifar10-cnn-med-jan04-1-5
                                           &                                      & SABR        & 54.20                               & 0.90                                    & 48.04                                             & 0.49  & \bf{39.96} & 0.30 & \bf{40.30} \\ % ./results/cifar10-cnn-med-jan04-1-5
                                           &                                      & Set (ours)  & 63.87                               & 2.23                                    & 55.17                                             & 1.56  & 37.79      & 1.55 & 39.36      \\ % ./tars/results/cifar10-cnn-med-set-jan05-1-5; tau=0.005
    \midrule
    % [top-3 of 5] TRADES, SABR, Set: ./results/svhn-cnn-med-dec31-1-5, IBP: ./tars/results/svhn-cnn-med-ibp-set-dec18-2217-1-5/,  (no grad clipping)
    \multirow{4}{*}{\textsc{Svnh}}         & \multirow{4}{*}{$0.01$}              & TRADES      & \bf{91.30}                          & 0.28                                    & \bf{78.34}                                        & 0.28  & 0.00       & 0.00 & 0.00       \\
                                           &                                      & IBP         & 72.33                               & 4.26                                    & 61.27                                             & 3.46  & \bf{46.93} & 3.09 & \bf{50.18} \\
                                           &                                      & SABR        & 81.79                               & 4.36                                    & 63.26                                             & 14.32 & 12.62      & 6.63 & 17.81      \\
                                           &                                      & Set (ours)  & 82.66                               & 3.13                                    & 72.35                                             & 2.68  & 39.60      & 2.39 & 41.35      \\
    \midrule
    % [top-3 of 5] Set: ./tars/results/tinyimagenet-cnn-med-set-dec21-2225, TRADES, IBP, SABR: ./results/250103-211009
    \multirow{4}{*}{\textsc{TinyImageNet}} & \multirow{4}{*}{$\nicefrac{1}{255}$} & TRADES      & \bf{28.14}                          & 0.56                                    & \bf{19.80}                                        & 0.43  & 0.00       & 0.00 & 0.00       \\
                                           &                                      & IBP         & 9.62                                & 1.17                                    & 8.76                                              & 0.95  & 2.89       & 1.27 & 4.07       \\ % + ./results/250106-180156 (run 4&5)
                                           &                                      & SABR        & 13.74                               & 1.38                                    & 12.49                                             & 1.25  & \bf{3.68}  & 0.87 & \bf{4.68}  \\ % + ./results/250107-180022 (run 4&5)
                                           &                                      & Set (ours)  & 16.70                               & 2.79                                    & 14.80                                             & 2.54  & 0.02       & 0.02 & 0.05       \\ % tau=0.1, ./case/results/tinyimagenet-cnn-med-set-0.1
    \bottomrule
  \end{tabular}
\end{table}

\cref{tab:main_results_table} shows our results.
Across all datasets, TRADES shows the best performance for clean and falsified accuracy; however, the verification is notoriously hard. Both interval-based approaches, IBP and SABR, have lower clean and falsified accuracy than TRADES but are significantly easier to verify. SABR achieves slightly better performance due to less regularization by propagating smaller intervals. For \textsc{SVHN}, SABR has lower verified accuracy than IBP, indicating that stronger verification algorithms are required. Our set-based training strikes a balance between TRADES and interval-based approaches, with significantly higher verified accuracy compared to TRADES and higher clean and falsified accuracy compared to IBP and SABR.

\subsection{Understanding Robustness}
\begin{figure}
  \centering
  \includetikz{decision-bounds}
  \caption{Comparing the decision bounds of point-based (left) and set-based (right) training. The dashed line is the decision boundary of point-based training.}\label{fig:decision-bounds}
\end{figure}
To better understand the robustness of neural networks, we compare the learned decision boundaries of standard (point-based) and set-based training for a simple binary classification task~(\cref{fig:decision-bounds}). Both training methods learn the training data perfectly, but set-based training pushes the decision boundaries away from the samples, making the set-based-trained model more robust. For some samples, the decision boundary of the point-based trained neural network crosses their perturbation sets, which is not the case for the set-based-trained neural network.

\subsection{Ablation Studies}
We conduct ablation studies to justify our design choices. Please refer to~\cref{sec:app:train-details} for details.
\paragraph{Weighting of Robustness Loss} \changed{It has been shown that there is a fundamental tradeoff between accuracy and robustness~\citep{zhang_et_al_2019}.} The weighting factor $\tau$ in our set-based loss~(\cref{def:set_loss_fun}) can trade off accuracy with robustness. \cref{tab:tau_compare} shows the accuracies for different values of $\tau$: For smaller values of $\tau$, the clean accuracy is larger, and the verified accuracy is larger for larger values. Thus, this hyperparameter can be used to tune the tradeoff between accuracy and robustness.
\begin{table}[tp]
  % \footnotesize
  \centering
  \caption{Performance on \textsc{Mnist} for different weighting factors in the set-based loss~(\cref{def:set_loss_fun}).}\label{tab:tau_compare}
  \begin{tabular}{l >{$}r<{$}@{$\,\pm\,$}>{$}l<{$} >{$}r<{$}@{$\,\pm\,$}>{$}l<{$} >{$}r<{$}@{$\,\pm\,$}>{$}l<{$} >{($}l<{$)}}
    \toprule
    \bf{Method}       & \multicolumn{2}{c}{\bf{clean Acc.}} & \multicolumn{2}{c}{\bf{falsified Acc.}} & \multicolumn{3}{c}{\bf{fast-verified Acc. (max)}}                                         \\ \midrule
    % [top-3 of 5] ./results/241213-110236/, cnn-tiny-bn, fgsm2-10+10, nn-verify(timeout=1s)
    Set ($\tau=0.0$)  & \bf{98.87}                          & 0.01                                    & 96.48                                             & 0.18 & 89.40      & 0.58 & 90.04      \\
    Set ($\tau=0.01$) & 98.74                               & 0.07                                    & \bf{96.77}                                        & 0.29 & 93.45      & 0.47 & 93.73      \\
    Set ($\tau=0.1$)  & 97.57                               & 0.09                                    & 95.60                                             & 0.17 & \bf{94.03} & 0.04 & \bf{94.06} \\
    %  from ./results/241220-095625/evaluation; Why so much better? 0.1       & 98.67                               & 0.13                                    & 97.17                                             & 0.19 & 96.39      & 0.09 & 96.50      \\
    Set ($\tau=0.2$)  & 97.17                               & 0.05                                    & 95.07                                             & 0.12 & 93.70      & 0.12 & 93.82      \\
    Set ($\tau=0.3$)  & 96.59                               & 0.07                                    & 93.46                                             & 1.20 & 88.91      & 4.36 & 93.95      \\ % \midrule
    % % ./results/241213-110236/, cnn-tiny-bn, fgsm2-10+10, nn-verify(timeout=0s)
    % 0.0       & \bf{98.86}                          & 0.02                                    & 96.24                                             & 0.34 & 18.69      & 12.20 & 32.78      \\
    % 0.01      & 98.60                               & 0.20                                    & \bf{96.41}                                        & 0.54 & 63.81      & 14.62 & 89.02      \\
    % 0.1       & 97.60                               & 0.11                                    & 95.64                                             & 0.15 & \bf{87.43} & 6.35  & 93.32      \\
    % 0.2       & 96.97                               & 0.27                                    & 94.40                                             & 1.54 & 87.40      & 10.12 & 93.47      \\
    % 0.3       & 96.56                               & 0.09                                    & 92.86                                             & 1.17 & 80.68      & 9.84  & \bf{93.85} \\
    \bottomrule
  \end{tabular}
\end{table}
\paragraph{Input Set} Compared to other robust training approaches~\citep{gowal_et_al_2019,mueller_et_al_2023,mao_et_al_2023}, our set-based training is not limited to multi-dimensional intervals as input sets. In \cref{tab:input_set_compare}, we compare using the $\ell_\infty$-ball as an input set with a smaller input zonotope where the generators are computed using adversarial attacks (see~\cref{sec:app:train-details} for details). The smaller input set leads to better performance; thereby, we confirm the observations from previous works~\citep{mueller_et_al_2023} that smaller input sets reduce the regularization by creating smaller approximation errors and thus resulting in better performance.
\begin{table}[tp]
  % \footnotesize
  \centering
  \caption{Performance on \textsc{Mnist} of different input sets.}\label{tab:input_set_compare}
  \begin{tabular}{l >{$}r<{$}@{$\,\pm\,$}>{$}l<{$} >{$}r<{$}@{$\,\pm\,$}>{$}l<{$} >{($}l<{$)}}
      \toprule
    \bf{Method}         & \multicolumn{2}{c}{\bf{clean Acc.}} & \multicolumn{3}{c}{\bf{fast-verified Acc. (max)}}                                  \\ \midrule
    Set ($\ell_\infty$) & \bf{97.58}                          & 0.03                                              & 93.83      & 0.06 & 93.89      \\ % Set (Zonotope+ours), Table 4
    Set (FGSM)          & 97.57                               & 0.09                                              & \bf{94.03} & 0.04 & \bf{94.06} \\ % Set ($\tau=0.1$), Table 2
    \bottomrule
  \end{tabular}
\end{table}
\changed{\paragraph{Set-Propagation Method} We compare the propagation of zonotopes with our image enclosure to other set propagation methods during our set-based training (\cref{tab:set_prop_compare}): IBP, zonotopes with Singh's enclosure, and zonotopes with our enclosure~(\cref{prop:set_forward_prop}). The results support our choice of enclosure as it consistently obtains higher accuracies than the other enclosures.
  This is due to the smaller accumulated approximation errors~(\cref{thm:img_enc_area_smaller}). Thus, we observe that our enclosure produces the best results, justifying the use of zonotopes for our image enclosure.}
\begin{table}[tp] % ran locally: ./results/241211-181518/, cnn-tiny-bn, l_inf-100; [other attempts: results/241210-172350, cnn-small-bn, fgsm-10-10]; verified with the corresponding method
  \centering
  \changed{
    \caption{Performance on \textsc{Mnist} of different set propagation methods.}\label{tab:set_prop_compare}
    \begin{tabular}{l >{$}r<{$}@{$\,\pm\,$}>{$}l<{$} >{$}r<{$}@{$\,\pm\,$}>{$}l<{$} >{$}r<{$}@{$\,\pm\,$}>{$}l<{$} >{($}l<{$)}}
      \toprule
      \bf{Method}          & \multicolumn{2}{c}{\bf{clean Acc.}} & \multicolumn{2}{c}{\bf{falsified Acc.}} & \multicolumn{3}{c}{\bf{fast-verified Acc. (max)}}                                         \\ \midrule
      % Set (Zonotope+ours)  & 97.58                               & 0.03                                    & \bf{95.31}                                  & 0.05 & \bf{93.83} & 0.06 \\
      % Set (Zonotope+Singh) & \bf{97.60}                          & 0.07                                    & 95.21                                       & 0.13 & 93.53      & 0.40 \\
      % Set (IBP)            & 93.16                               & 0.16                                    & 74.90                                       & 1.04 & 0.15       & 0.17 \\
      % ./results/mnist-singh-ablation
      Set (Zonotope+ours)  & \bf{98.60}                          & 0.21                                    & \bf{97.13}                                        & 0.13 & \bf{96.13} & 0.13 & \bf{96.25} \\
      Set (Zonotope+Singh) & 98.36                               & 0.13                                    & 96.60                                             & 0.59 & 94.44      & 2.57 & 96.04      \\
      % ./results/mnist-singh-ablation-ival
      Set (IBP)            & 97.77                               & 0.19                                    & 96.11                                             & 0.20 & 95.35      & 0.27 & 95.62      \\
      \bottomrule
    \end{tabular}
  }
\end{table}

\changed{
  \paragraph{Size of the Output Sets and Robustness} In~\cref{tab:interval_norm_comp}, we compare the size of the output sets (of the first 1000 test samples) and the Lipschitz constant of the best performing neural networks trained on \textsc{Mnist}.
  The Lipschitz constant of a neural network is another metric for robustness, because it bounds the sensitivity of the neural network for input changes~\citep{fazlyab_et_al_2019}. Moreover, we compare the interval norm~\citep[Sec.~3.1]{althoff_2023} of the output set, because computing the volume of a zonotope is computationally hard. The Lipschiz constant of the TRADES-trained network is the smallest, which explains the great empirical robustness (falsified Acc. in~\cref{tab:main_results_table}); however, the output sets are too large for verification (fast-verified Acc. in~\cref{tab:main_results_table}). Our set-based-trained neural network produces the smallest output sets and a smaller Lipschitz constant compared to IBP and SABR.
  \begin{table}[t]
    \centering
    \changed{
      \caption{\changed{Comparing the sizes of the output sets and Lipschitz for different training methods.}}\label{tab:interval_norm_comp}
      \begin{tabular}{l >{$}r<{$}@{$\,\pm\,$}>{$}l<{$} >{$}c<{$}}
        \toprule
        \bf{Method} & \multicolumn{2}{c}{\bf{Size (Interval Norm)}} & \text{\bf{Lipschitz Constant}}                          \\ \midrule
        TRADES      & 276.36                                        & 120.40                         & \bf{9.72\times 10^{7}} \\
        IBP         & 4.77                                          & 1.96                           & 2.78\times 10^{16}     \\
        SABR        & 5.26                                          & 2.39                           & 4.20\times 10^{17}     \\
        Set (ours)  & \bf{1.85}                                     & 1.27                           & 3.49\times 10^{12}     \\
        \bottomrule
      \end{tabular}
    }
  \end{table}
}

% !TeX root = ../paper.tex
% Add the above to each tex file to make compiling the PDF easier in some editors.

\section{Conclusion}\label{sec:concl}

This paper introduces the first set-based training procedure for neural networks that uses gradient sets: During training, we enclose the output set of the neural network using set propagation and derive a gradient set, which contains a different gradient for each possible output. By choosing gradients that point to the center of the output set, we can directly reduce the size of the output set. Thereby, we can simultaneously improve robustness and simplify the formal verification.
The set-based training is made possible by a fast, batch-wise, and differentiable propagation of zonotopes, which utilizes analytical solutions for approximation errors.
Our experimental results demonstrate that our set-based approach effectively trains robust neural networks, which have competitive performance and admit fast verification (in polynomial time). Thereby, we demonstrate that gradient sets can be effectively used to train robust neural networks. Hence, set-based training represents a promising new direction for robust neural network training and a significant step toward fast verification of neural networks and, thus, the widespread adoption of formal verification of neural networks.

\section*{Acknowledgements}
\addcontentsline{toc}{section}{Acknowledgment}

This work was partially supported by the project SPP-2422 (No. 500936349) and the project FAI (No. 286525601), both funded by the German Research Foundation (DFG). We also want to thank our colleague Mark Wetzlinger from our research group for his revisions of the manuscript.

\bibliography{refs-clean-short}
\bibliographystyle{tmlr}

\appendix
\crefalias{section}{appendix}

\section{Evaluation Details}\label{sec:app:train-details}

\paragraph{Hardware}

Our experiments were run on a server with $2\times$AMD EPYC $7763$ ($64$ cores/$128$ threads), $2$TB RAM, and a NVIDIA A$100$ $40$GB GPU.

\paragraph{Training Hyperparameters}

The training hyperparameters are listed in~\cref{tab:hyperparams} and the neural network architectures in~\cref{tab:network_arch}. We use the same notation as~\citep{gowal_et_al_2019}: $\textsc{Conv}\,k\;w\times h + s$ denotes a convolutional layer with $k$ filters of size $w\times h$ and stide $s$, and $\textsc{Fc}\,n$ denotes a fully connected layer with $n$ neurons.
The weights and biases are initialized as in~\citep{shi_et_al_2021}. We use Adam optimizer~\citep{kingma_ba_2015} with the recommended hyperparameters.
For all training methods, we tried to use hyperparameters as close as possible to the reported hyperparameters in their respective paper: for \citep{gowal_et_al_2019} we use $\kappa=\nicefrac{1}{2}$, for \citep{zhang_et_al_2019} we use $\nicefrac{1}{\lambda}=6$, for \citep{mueller_et_al_2023} we use $\lambda=0.1$ for \textsc{Cifar-10}, and $\lambda=0.4$ for \textsc{Mnist} and \textsc{Svhn}.
For any PGD during training (used by methods \citep{madry_et_al_2018,zhang_et_al_2019,mueller_et_al_2023}) we used the settings from \citep{mueller_et_al_2023}: $8$ iterations with an initial step size $0.5$, which is decayed twice by $0.1$ at iterations $4$ and $7$. All PGD attacks for testing are computed with $40$ iterations of step size $0.01$.
Moreover, all reported accuracies are averaged over the best $3$ of $5$ runs. We clip gradients with a $\ell_2$-norm of greater than $10$. All reported perturbation radii are w.r.t. normalized inputs between $0$ and $1$. To reduce computational resources, we performed the ablation studies with a smaller 3-layer convolutional neural network trained on \textsc{Mnist}.

\begin{table*}
  \centering
  \caption{Training hyperparameters.}\label{tab:hyperparams}
  \begin{tabular}{ l >{$}c<{$} >{$}c<{$} >{$}c<{$} >{$}c<{$} >{$}c<{$} >{$}c<{$}}
    \toprule
                          &                &                   &           &                     & \textbf{\#Epochs}                            &                \\
    \bf{Dataset}          & \bf{\eta}      & \bf{\epsilon}     & \bf{\tau} & \textbf{Batch Size} & \text{(\textit{warm-up} / \textit{ramp-up})} & \textbf{Decay} \\ \midrule
    \textsc{Mnist}        & 5\cdot 10^{-4} & 0.1               & 0.1       & 256                 & 70\, (1 / 20)                                & 50, 60         \\
    \textsc{Cifar-10}     & 5\cdot 10^{-4} & \nicefrac{2}{255} & 0.005     & 128                 & 160\, (40 / 120)                             & 120, 140       \\
    \textsc{Svhn}         & 5\cdot 10^{-4} & 0.01              & 0.01      & 128                 & 70\, (20 / 50)                               & 50, 60         \\
    \textsc{TinyImageNet} & 5\cdot 10^{-4} & \nicefrac{1}{255} & 0.1       & 64                  & 70\, (20 / 50)                               & 50, 60         \\
    \bottomrule
  \end{tabular}
\end{table*}

\begin{table}
  \centering
  \caption{Neural network architectures. Each linear layer or convolutional layer (except the last layer) is followed by a batch normalization and a nonlinear activation layer ($\relu$).}\label{tab:network_arch}
  \begin{tabular}{ >{$}c<{$} >{$}c<{$} }
    \toprule
    \bf{\textsc{Cnn3}}                 & \bf{\textsc{Cnn6}}                 \\ \midrule % \bf{\textsc{Cnn3}}                
    \textsc{Conv}\, 5\; 4\times 4 + 2  & \textsc{Conv}\, 32\; 3\times 3 + 1 \\          % \textsc{Conv}\, 32\; 4\times 4 + 2 
    \textsc{Conv}\, 10\; 4\times 4 + 1 & \textsc{Conv}\, 32\; 4\times 4 + 2 \\          % \textsc{Conv}\, 64\; 4\times 4 + 1
    \textsc{Fc}\, 100                  & \textsc{Conv}\, 64\; 3\times 3 + 1 \\          % \textsc{Fc}\, 100                 
                                       & \textsc{Conv}\, 64\; 4\times 4 + 2 \\
                                       & \textsc{Fc}\, 512                  \\
                                       & \textsc{Fc}\, 512                  \\
    \bottomrule
  \end{tabular}
\end{table}

\paragraph{Datasets}

\textsc{Mnist} contains $60\,000$ grayscale images of size $28\times 28$. Each image depicts a handwritten digit from $0$ to $9$. \textsc{Svhn} is a real-world dataset that contains $73\,257$ colored images of digits of house numbers that are cropped to size $32\times 32$. The \textsc{Cifar-10} dataset contains $60\,000$ colored images of size $32\times 32$. The \textsc{TinyImageNet} data set contains $100\,000$ colored images of size $64\times 64$ labeled with $200$ classes. We use the canonical split of training and test data for each dataset and the entire test data for evaluation; because test labels are not available for \textsc{TinyImageNet}, we follow~\citep{mueller_et_al_2023} and use the validation set for testing.
Following~\citep{xu_et_al_2020}, we augment the \textsc{Cifar-10} and \textsc{TinyImageNet} dataset with random crop and flips.
The perturbation is applied before the normalization to ensure comparability with the literature.

\paragraph{Improvements for Scalability}

The complexity of our set-based training depends on the number of generators of the output set used during training~(\cref{prop:set_training_complexity}). We use two methods to reduce the number of generators:
\begin{enumerate*}[label=(\roman*)]
  \item We propagate the approximation errors as intervals through the neural network and compute the Minkowski sum at the end.
  \item We use input sets with fewer number of generators.
\end{enumerate*}
Our set-based training can use any zonotopic input set and is not limited to $\ell_\infty$-input sets. Building on previous work~\citep{mueller_et_al_2023}, we use adversarial attacks to construct smaller input sets that focus on critical regions of the input; thereby, the regularization through large approximation errors is reduced. Previous works only use a single adversarial attack to construct a smaller $\ell_\infty$-input set~\citep{mueller_et_al_2023}. We extend this idea and use several adversarial attacks to construct zonotopic input sets: We shift the center of the input set to the average attack and use scaled directions of the attacks as generators for the input set. Given an input $\nnInput\in\R^{\numNeurons_0}$, we compute $\lambda$ adversarial attacks $\tilde{\nnInput}_i = \nnInput + \delta_i$ for $i\in[\lambda]$ (e.g., using FGSM). The input set is constructed as
\begin{equation*}
  \nnInputSet=\shortZ[.]{\nnInput + \nicefrac{1}{\lambda}\,\sum_{\lambda}^{i=1}\delta_i}{\nicefrac{1}{\lambda}\,\cmat{\delta_1 & \delta_2 & \cdots & \delta_\lambda}}
\end{equation*}

\paragraph{Fairness}

We note that compared to the literature, we use smaller neural networks and compare the means across several training runs; most literature only reports their best-observed scores, which is problematic regarding reproducibility and fairness because it is unclear how many training seeds were used. Therefore, to create a fair comparison, we have reimplemented related approaches~\citep{madry_et_al_2018,gowal_et_al_2019,zhang_et_al_2019,mueller_et_al_2023}.
The implementations have been validated by reproducing their reported results.

\changed{
  \section{Additional Experiments}\label{app:sec:add_experiments}

  \paragraph{Maximum Perturbation Radius} In~\cref{tab:max_epsilon_comp}, we compare the maximal perturbation radius for which we can verify the robustness. We used binary search (10 iterations) to compute the maximum perturbation radius for the first 1000 test samples of the \textsc{Mnist} dataset. The perturbation radii of IBP, SABR, and Set are comparable and significantly higher compared to TRADES.
  \begin{table}[t]
    \centering
    \changed{
      \caption{Comparing the maximum perturbation radius.}\label{tab:max_epsilon_comp}
      \begin{tabular}{l >{$}r<{$}@{$\,\pm\,$}>{$}l<{$} >{($}l<{$)}}
        \toprule
        \bf{Method} & \multicolumn{3}{c}{\bf{$\epsilon$ (max)}}                        \\ \midrule
        TRADES      & 0.0715                                    & 0.0140 & 0.1094      \\
        IBP         & \bf{0.1856}                               & 0.0528 & 0.2441      \\
        SABR        & 0.1792                                    & 0.0482 & 0.2422      \\
        Set (ours)  & 0.1782                                    & 0.0494 & \bf{0.2676} \\
        \bottomrule
      \end{tabular}
    }
  \end{table}

  \paragraph{Weighting of Robustness Loss} \cref{tab:tau_compare_cifar10} shows the accuracies for different values of $\tau$ of \textsc{Cnn3} trained on the \textsc{Cifar10} dataset.
  \begin{table}
    % \footnotesize
    \centering
    \changed{
      \caption{Performance on \textsc{Cifar10} for different weighting factors in the set-based loss~(\cref{def:set_loss_fun}).}\label{tab:tau_compare_cifar10}
      \begin{tabular}{l >{$}r<{$}@{$\,\pm\,$}>{$}l<{$} >{$}r<{$}@{$\,\pm\,$}>{$}l<{$} >{$}r<{$}@{$\,\pm\,$}>{$}l<{$} >{($}l<{$)}}
        \toprule
        \bf{Method}        & \multicolumn{2}{c}{\bf{clean Acc.}} & \multicolumn{2}{c}{\bf{falsified Acc.}} & \multicolumn{3}{c}{\bf{fast-verified Acc. (max)}}                                         \\ \midrule
        % [top-3 of 5] ./results/cifar10-tau-ablation/
        Set ($\tau=0.0$)   & \bf{61.42}                          & 1.12                                    & 50.80                                             & 0.62 & 31.98      & 0.61 & 32.53      \\
        Set ($\tau=0.005$) & 59.95                               & 1.13                                    & \bf{50.89}                                        & 0.64 & 38.50      & 0.32 & 38.87      \\
        Set ($\tau=0.01$)  & 59.29                               & 0.99                                    & 50.37                                             & 0.76 & 39.59      & 0.31 & 39.92      \\
        %  from ./results/cifar10-tau-ablation-2/
        Set ($\tau=0.05$)  & 56.75                               & 0.79                                    & 48.80                                             & 0.46 & 41.10      & 0.13 & 41.24      \\
        Set ($\tau=0.1$)   & 54.90                               & 0.84                                    & 47.90                                             & 0.58 & \bf{41.57} & 0.28 & \bf{41.87} \\
        %  from ./results/cifar10-tau-ablation-3/
        Set ($\tau=0.2$)   & 50.53                               & 1.18                                    & 45.05                                             & 0.83 & 40.08      & 0.58 & 40.49      \\
        Set ($\tau=0.3$)   & 45.55                               & 1.77                                    & 41.29                                             & 1.51 & 36.13      & 2.97 & 38.23      \\
        \bottomrule
      \end{tabular}
    }
  \end{table}

  \changed{\paragraph{Set-Propagation Method} \cref{tab:set_prop_compare_cnn6} shows the results of \textsc{Cnn6} trained on \textsc{Mnist} for different set propagation methods.}
  \begin{table}[tp]
    \centering
    \changed{
      \caption{Performance of \textsc{Cnn6} on \textsc{Mnist} of different set propagation methods.}\label{tab:set_prop_compare_cnn6}
      \begin{tabular}{l >{$}r<{$}@{$\,\pm\,$}>{$}l<{$} >{$}r<{$}@{$\,\pm\,$}>{$}l<{$} >{$}r<{$}@{$\,\pm\,$}>{$}l<{$} >{($}l<{$)}}
        \toprule
        \bf{Method}          & \multicolumn{2}{c}{\bf{clean Acc.}} & \multicolumn{2}{c}{\bf{falsified Acc.}} & \multicolumn{3}{c}{\bf{fast-verified Acc. (max)}}                                         \\ \midrule
        % ./results/mnist-singh-ablation
        Set (Zonotope+ours)  & \bf{98.66}                          & 0.02                                    & \bf{97.71}                                        & 0.08 & \bf{97.08} & 0.08 & \bf{97.15} \\
        Set (Zonotope+Singh) & 98.13                               & 0.88                                    & 97.61                                             & 0.19 & 96.88      & 0.19 & 97.08      \\
        Set (IBP)            & 98.08                               & 0.08                                    & 96.60                                             & 0.21 & 95.96      & 0.21 & 96.11      \\
        \bottomrule
      \end{tabular}
    }
  \end{table}

  % \paragraph{Set-Propagation Method (\textsc{Cnn6})} We compare the effect of different output enclosures on our set-based training with \textsc{Cnn6} trained on the \textsc{Mnist} dataset.
  % \begin{table}[t]
  %   \centering
  %   \caption{Performance of \textsc{Cnn6} on \textsc{Mnist} of different set propagation methods.}\label{tab:set_prop_compare_cnn6}
  %   \begin{tabular}{>{\color{blue}}l >{\color{blue}$}r<{$}@{\color{blue}$\,\pm\,$}>{\color{blue}$}l<{$} >{\color{blue}$}r<{$}@{\color{blue}$\,\pm\,$}>{\color{blue}$}l<{$} >{\color{blue}$}r<{$}@{\color{blue}$\,\pm\,$}>{\color{blue}$}l<{$}}
  %     \toprule
  %     \bf{Method}          & \multicolumn{2}{c}{\color{blue}\bf{clean Acc.}} & \multicolumn{2}{c}{\color{blue}\bf{falsified Acc.}} & \multicolumn{2}{c}{\color{blue}\bf{fast-verified Acc.}}         \\ \midrule
  %     % ./results/mnist-singh-ablation-cnn6
  %     Set (Zonotope+ours)  &                                                 &                                                     &                                                         &  &  & \\
  %     Set (Zonotope+Singh) &                                                 &                                                     &                                                         &  &  & \\
  %     % ./results/
  %     Set (IBP)            &                                                 &                                                     &                                                         &  &  & \\
  %     \bottomrule
  %   \end{tabular}
  % \end{table}
}

\paragraph{Training Times}
The training times are compared in~\cref{tab:train_time}. \changed{The computational overhead of set-based training is slightly higher than TRADES, IBP, and SABR.
  However, the TRADES-trained neural networks have great empirical performance (clean and falsified accuracy~\cref{tab:main_results_table}) but cannot be verified using polynomial time verification approaches (fast-verified accuracy~\cref{tab:main_results_table}).
  IBP and SABR have significantly lower empirical performance compared to TRADES.
  Our set-based trained neural networks have higher empirical performance compared to IBP and SABR and have great verified robustness (fast-verified accuracy~\cref{tab:main_results_table}).
  Moreover, with our set-based training, we can use a weighting parameter in the set-based loss to explicitly tune the tradeoff between empirical performance and verified robustness.
  Therefore, set-based training strikes a balance between training time, empirical performance, and verified robustness.}

\begin{table}[t]
  \centering
  \caption{Training time with \textsc{Cnn6} on \textsc{Mnist} (min. of 5 runs) [sec / Epoch].}\label{tab:train_time}%
  \begin{tabular}{l >{$}c<{$}}
    \toprule
    \bf{Method} & \textbf{Training Time [sec / Epoch]} (\downarrow) \\ \midrule
    Point       & 6.1                                               \\
    TRADES      & 49.6                                              \\
    IBP         & 19.9                                              \\
    SABR        & 52.0                                              \\
    Set (ours)  & 61.2                                              \\
    \bottomrule
  \end{tabular}
\end{table}

\section{Reproducibility of \cref{fig:decision-bounds}}

\newcommand{\vecTwot}[2]{\begin{bsmallmatrix} {\displaystyle #1} && {\displaystyle #2} \end{bsmallmatrix}^\top}
\newcommand{\vecTwo}[2]{\begin{bsmallmatrix} {\displaystyle #1} \\ {\displaystyle #2} \end{bsmallmatrix}}

\cref{fig:decision-bounds} compares the decision boundaries of a point-based and a set-based training of a neural network for a binary classification task. The network architecture is nn-med: $5$ layers with $100$ neurons each. The training data are $20$ random input samples $\nnInput_i\in[0,1]^2$ with corresponding targets $\nnTarget_i\in\{0,1\}^2$:
\begin{align*}
  \mathcal{D}=\Bigl\{
   & \left(\vecTwo{0.0622}{0.6995},\vecTwo{0}{1}\right),\left(\vecTwo{0.6534}{0.9409},\vecTwo{0}{1}\right),\left(\vecTwo{0.4759}{0.7163},\vecTwo{1}{0}\right),\left(\vecTwo{0.8812}{0.1020},\vecTwo{1}{0}\right),\left(\vecTwo{0.5047}{0.4685},\vecTwo{1}{0}\right), \\
   & \left(\vecTwo{0.1470}{0.3275},\vecTwo{1}{0}\right),\left(\vecTwo{0.3439}{0.1395},\vecTwo{0}{1}\right),\left(\vecTwo{0.9098}{0.5422},\vecTwo{0}{1}\right),\left(\vecTwo{0.8588}{0.8696},\vecTwo{0}{1}\right),\left(\vecTwo{0.0545}{0.0825},\vecTwo{1}{0}\right), \\
   & \left(\vecTwo{0.6889}{0.4771},\vecTwo{1}{0}\right),\left(\vecTwo{0.9329}{0.2857},\vecTwo{0}{1}\right),\left(\vecTwo{0.6781}{0.3043},\vecTwo{0}{1}\right),\left(\vecTwo{0.4641}{0.3302},\vecTwo{0}{1}\right),\left(\vecTwo{0.4575}{0.9487},\vecTwo{0}{1}\right), \\
   & \left(\vecTwo{0.1272}{0.4699},\vecTwo{1}{0}\right),\left(\vecTwo{0.6506}{0.7315},\vecTwo{0}{1}\right),\left(\vecTwo{0.5207}{0.1229},\vecTwo{1}{0}\right),\left(\vecTwo{0.3271}{0.4574},\vecTwo{1}{0}\right),\left(\vecTwo{0.6858}{0.0616},\vecTwo{1}{0}\right)
  \Bigr\}\text{.}
\end{align*}
We train both neural networks for $200$ epochs with a mini-batch size of $10$ using the Adam optimizer with a learning rate of $\eta=0.01$. For set-based training, we use $\epsilon=0.05$ with $\tau=0.1$.

% \section{Reproducibility of~\cref{fig:img_enc_output_sets}}

% \cref{fig:img_enc_output_sets} compares the output sets computed with our image enclosure and Singh's enclosure. We use the point-based trained neural network from our first training run of nn-med on MNIST. The depicted output sets are computed for the $8$-th image of the test set with a perturbation radius $\epsilon=0.005$.

\section{Proofs}\label{app:sec:proofs}

\begin{repproposition}{prop:grad_frad_norm}
  \propGradFradNorm
  \begin{proof}
    The center does not affect the F-radius, hence $\grad{c_\numLayers}{\norm{\nnOutputSet}_{F}} = \zeros$.
    The F-radius is the sum of all squared entries of the generator matrix. Hence,
    \begin{align}\label{eq1:prop:grad_frad_norm}
      \grad{G_{\numLayers}}{\norm{\nnOutputSet}_{F}} = \frac{1}{\numNeurons_\numLayers}\,\grad{G_{\numLayers}}{\sqrt{\ones^\top\,(G_\numLayers\odot G_\numLayers)\,\ones}} = \frac{G_{\numLayers}}{\numNeurons_\numLayers\,\norm{\nnOutputSet}_{F}}\text{.}
    \end{align}
    Thus,
    \begin{align*}
      \grad{\nnOutputSet}{\norm{\nnOutputSet}_{F}} \overset{\text{\cref{def:zonotope_gradient}}} & {=} \shortZ{\grad{c_\numLayers}{\norm{\nnOutputSet}_{F}}}{\grad{G_\numLayers}{\norm{\nnOutputSet}_{F}}} \overset{\text{\cref{eq1:prop:grad_frad_norm}}}{=} \shortZ{\zeros}{\frac{G_\numLayers}{\numNeurons_\numLayers\,\norm{\nnOutputSet}_{F}}} = \frac{1}{\numNeurons_\numLayers\,\norm{\nnOutputSet}_{F}}\,\shortZ{\zeros}{G_\numLayers}\text{.}\qedhere
    \end{align*}
  \end{proof}
\end{repproposition}

\begin{repproposition}{prop:grad_set_loss}[Set-Based Loss Gradient]
  \propGradSetLoss{}
  \begin{proof}
    This follows from~\cref{def:set_loss_fun,prop:grad_frad_norm}:
    \begin{align*}
      \grad{\nnOutputSet}{\setLoss(\nnTarget,\nnOutputSet)} \overset{\text{\cref{def:set_loss_fun}}} & {=} (1 - \tau)\,\grad{\nnOutputSet}{\pointLoss(\nnTarget,c_\numLayers)} + \frac{\tau}{\epsilon}\,\grad{\nnOutputSet}{\norm{\nnOutputSet}_F} \overset{\text{\cref{def:zonotope_gradient}}} {=} (1 - \tau)\,\shortZ[]{\grad{c_\numLayers}{\pointLoss(\nnTarget,c_\numLayers)}}{\zeros} + \frac{\tau}{\epsilon}\,\grad{\nnOutputSet}{\norm{\nnOutputSet}_F} \\
      \overset{\text{\cref{prop:grad_frad_norm}}}                                                    & {=} (1 - \tau)\,\shortZ[]{\grad{c_\numLayers}{\pointLoss(\nnTarget,c_\numLayers)}}{\zeros} + \frac{\tau}{\epsilon\,\numNeurons_\numLayers\,\norm{\nnOutputSet}_{F}}\,\shortZ{\zeros}{G_\numLayers}                                                                                                                                                       \\
                                                                                                     & = \shortZ[.]{(1 - \tau)\,\grad{c_\numLayers}{\pointLoss(\nnTarget,c_\numLayers)}}{\frac{\tau}{\epsilon\,\numNeurons_\numLayers\,\norm{\nnOutputSet}_{F}}\,G_\numLayers}\qedhere
    \end{align*}
  \end{proof}
\end{repproposition}

\begin{repproposition}{prop:approx_error}
  \propApproxError
  \begin{proof}
    \begin{enumerate}[label=Case (\roman*)., wide=0pt, font=\itshape]
      \item $\actfun = \relu$. $\relu(x) - \approxFun(x)$ is linear for $[l,0]$ and $[0,u]$. Thus, the approximation errors are found at the bounds $x\in\{l,u\}$ or where $0\in[l,u]$ at $x = 0$.
      \item $\actfun = \tanh$. The derivative of the hyperbolic tangent is $\tanh'(x) = 1 - \tanh(x)^2$. To compute the extreme points of $\tanh(x) - \approxFun(x)$, we demand that its derivative is $0$ and simplify the terms:
            \begin{align*}
                              &  & 0 & = \nicefrac{\mathrm{d}}{\mathrm{d} x} \left(\tanh(x) - \approxFun(x)\right) &
              \Leftrightarrow &  & 0 & = 1 - \tanh(x)^2 - \approxSlope                                             &
              \Leftrightarrow &  & x & = \pm\arctanh\left(\sqrt{1 - \approxSlope}\right)\text{.}
            \end{align*}
      \item $\actfun = \sigmoid$. To compute the extreme points of $\sigmoid(x) - \approxFun(x)$, we demand that its derivative is $0$ and simplify the terms:
            \begin{align*}
                              &  & 0 & = \nicefrac{\mathrm{d}}{\mathrm{d} x} (\sigmoid(x) - \approxFun(x))                                                                   &
              \Leftrightarrow &  & 0 & = \nicefrac{\mathrm{d}}{\mathrm{d} x}\left(\nicefrac{1}{2}\,\left(\tanh\left(\nicefrac{x}{2}\right) + 1\right) - \approxFun(x)\right) &
              \Leftrightarrow &  & x & = \pm 2\,\arctanh\left(\sqrt{1-4 \,\approxSlope}\right)\text{.}\qedhere
            \end{align*}
    \end{enumerate}
  \end{proof}
\end{repproposition}

\begin{figure}
  \centering
  \includetikz{prop_approx_error}
  \caption{Illustration for~\cref{thm:img_enc_area_smaller}.}\label{fig:prop_opt_approx_error}
\end{figure}

\begin{repproposition}{thm:img_enc_area_smaller}
  \thmImgEncAreaSmaller
  \begin{proof}
    We first observe that the approximation errors $\approxErrorL$ and $\approxErrorU$ of $\approxFun$ can be computed at points $\xapproxErrorU,\xapproxErrorL\in[l,u]$ such that $\xapproxErrorL\leq \xapproxErrorU$~(\cref{fig:prop_opt_approx_error}):
    \begin{align}\label{eq:thm1_eq1}
      \approxErrorL & = \actfun(\xapproxErrorL) - \approxFun(\xapproxErrorL)\text{,} & \approxErrorU & = \actfun(\xapproxErrorU) - \approxFun(\xapproxErrorU)\text{.}
    \end{align}
    Singh's enclosure~\citep[Thm.~3.2]{singh_et_al_2018} uses the linear approximation $\approxFun_{\singhSub}(x) = \approxSlope_{\singhSub}\, x + \approxOffset_{\singhSub}$, with slope $\approxSlope_{\singhSub}$, offset $\approxOffset_{\singhSub}$, and approximation error $\approxError_{\singhSub}$:
    \begin{align}\label{eq:singh_def}
      \approxSlope_{\singhSub}  & = \min\left(\actfun'(l),\actfun'(u)\right)\text{,}                                       &
      \approxOffset_{\singhSub} & = \nicefrac{1}{2}\,(\actfun(u) + \actfun(l) - \approxSlope_{\singhSub}\,(u + l))\text{,} &
      \approxError_{\singhSub}  & = \nicefrac{1}{2}\,(\actfun(u) - \actfun(l) - \approxSlope_{\singhSub}\,(u - l))\text{.}
    \end{align}
    Using \cref{def:lin_approx}, we obtain the following inequality:
    \begin{equation*}
      \approxSlope = \frac{\actfun\left(u\right) - \actfun\left(l\right)}{u - l} \geq \approxSlope_{\singhSub}\text{.}
    \end{equation*}
    Hence, we have
    \begin{equation}\label{eq:thm1_eq2}
      \approxSlope\,(\xapproxErrorU - \xapproxErrorL) \geq \approxSlope_{\singhSub}\,(\xapproxErrorU - \xapproxErrorL)\text{.}
    \end{equation}
    Moreover, from \cref{eq:singh_def} we have for all $x\in[l,u]$:
    \begin{align}
      \begin{split}\label{eq:thm1_eq3}
        \actfun(x) - \actfun(l) \geq \approxSlope_{\singhSub}\,(x - l) & \implies \actfun(\xapproxErrorL) \geq \actfun(l) + \approxSlope_{\singhSub}\,(\xapproxErrorL - l)\text{,}\\
        \actfun(u) - \actfun(x) \geq \approxSlope_{\singhSub}\,(u - x) & \implies \actfun(\xapproxErrorU) \leq \actfun(u) - \approxSlope_{\singhSub}\,(u - \xapproxErrorU)\text{.}
      \end{split}
    \end{align}
    Ultimately, we have
    \begin{align}\label{eq:dudl_leq_twods}
      \begin{split}
        \approxErrorU - \approxErrorL  \overset{\text{\cref{eq:thm1_eq1}}} & {=} \actfun(\xapproxErrorU) - \approxFun(\xapproxErrorU) - (\actfun(\xapproxErrorL) - \approxFun(\xapproxErrorL)) \overset{\text{\cref{def:lin_approx}}}{=} \actfun(\xapproxErrorU) - (\approxSlope\, \xapproxErrorU) - (\actfun(\xapproxErrorL) - (\approxSlope\, \xapproxErrorL)) \\
        \overset{\text{\cref{eq:thm1_eq2}}} & {\leq} \actfun(\xapproxErrorU) - \actfun(\xapproxErrorL) - \approxSlope_{\singhSub}\,(\xapproxErrorU - \xapproxErrorL) \overset{\text{\cref{eq:thm1_eq3}}}{\leq}
        \actfun(u) - \approxSlope_{\singhSub}\,(u - \xapproxErrorU) - (\actfun(l) + \approxSlope_{\singhSub}\,(\xapproxErrorL - l)) - \approxSlope_{\singhSub}\,(\xapproxErrorU - \xapproxErrorL) \overset{\text{\cref{eq:singh_def}}}{=} 2\, \approxError_{\singhSub}\text{.}
      \end{split}
    \end{align}
    Hence, we obtain the bound $\approxErrorU - \approxErrorL \leq 2\, \approxError_{\singhSub}$.
    Thus,
    \begin{align*}
      \area(\shortI{\approxErrorL}{\approxErrorU},[l,u]) \overset{\text{\cref{eq:img_enc_area}}} & {=} (u - l)\,(\approxErrorU - \approxErrorL) \overset{\text{\cref{eq:dudl_leq_twods}}}{\leq} (u - l)\, 2\, \approxError_{\singhSub} \overset{\text{\cref{eq:img_enc_area}}}{=} \area(\shortI{-\approxError_{\singhSub}}{\approxError_{\singhSub}},[l,u])\text{.}
    \end{align*}
  \end{proof}
\end{repproposition}

\begin{repproposition}{prop:backprop_fast_enclose}[Backpropagation through Image Enclosure]
  \propBackpropFastEnclose
  \begin{proof}
    The image enclosure adds $\numNeurons_k$ generators, hence the input set $\nnHiddenSet_{k-1}$ has $\numNeurons_k$ generators less than gradient set $\nnGradSet_{k}$, i.e. $G_{k-1}\in\R^{\numNeurons_k\times p}$ and $G'_{k}\in\R^{\numNeurons_k\times q}$ with $q = p + \numNeurons_k$.

    We unfold \cref{eq:set-backprop-grads} and rewrite the gradient set $\nnGradSet_{k-1}$ using the chain rule for partial derivatives.
    Let $\nnHiddenSet_{k} = \shortZ{c_{k}}{G_{k}}$ with $G_{k}\in\R^{\numNeurons_k\times q}$ be the output set of the $k$-th layer and let $\nnGradSet_{k} = \shortZ{c'_{k}}{G'_{k}}$ be the gradient set w.r.t. $\nnHiddenSet_{k}$:
    \begin{align}\label{eq:grad-chain-rule-part-der}
      \begin{split}
        \nnGradSet_{k-1} & \overset{\cref{eq:set-backprop-grads}}{=} \grad{\nnHiddenSet_{k-1}}{\setLoss(\nnTarget,\nnOutputSet)} = \sum_{i=1}^{\numNeurons_k} \partgrad{c_{k(i)}}{\setLoss(\nnTarget,\nnOutputSet)}\,\grad{\nnHiddenSet_{k-1}}{c_{k(i)}} + \sum_{i=1}^{\numNeurons_k}\sum_{j=1}^{q} \partgrad{G_{k(i,j)}}{\setLoss(\nnTarget,\nnOutputSet)}\,\grad{\nnHiddenSet_{k-1}}{G_{k(i,j)}} \\
        & = \sum_{i=1}^{\numNeurons_k} c'_{k(i)}\,\grad{\nnHiddenSet_{k-1}}{c_{k(i)}} + \sum_{i=1}^{\numNeurons_k}\sum_{j=1}^{q} G'_{k(i,j)}\,\grad{\nnHiddenSet_{k-1}}{G_{k(i,j)}}\text{.}
      \end{split}
    \end{align}

    We split \cref{eq:grad-chain-rule-part-der} into three summands:
    \begin{align*}
      \nnGradSet_{k-1,c} & = \sum_{i=1}^{\numNeurons_k} c'_{k(i)}\,\grad{\nnHiddenSet_{k-1}}{c_{k(i)}}\text{,}                     \\
      \nnGradSet_{k-1,p} & = \sum_{i=1}^{\numNeurons_k}\sum_{j=1}^{p} G'_{k(i,j)}\,\grad{\nnHiddenSet_{k-1}}{G_{k(i,j)}}\text{,}   \\
      \nnGradSet_{k-1,q} & = \sum_{i=1}^{\numNeurons_k}\sum_{j=p+1}^{q} G'_{k(i,j)}\,\grad{\nnHiddenSet_{k-1}}{G_{k(i,j)}}\text{.}
    \end{align*}
    Hence,
    \begin{align}\label{eq:prop:bpfastenc:1}
      \nnGradSet_{k-1} = \nnGradSet_{k-1,c} + \nnGradSet_{k-1,p} + \nnGradSet_{k-1,q}\text{.}
    \end{align}

    Furthermore, the input set $\nnHiddenSet_{k-1}$ is enclosed by the interval $\shortI{l_{k-1}}{u_{k-1}}$, where $l_{k-1} = c_{k-1} - \abs{G_{k-1}}\,\ones$ and $u_{k-1} = c_{k-1} + \abs{G_{k-1}}\,\ones$ (\cref{prop:interval_enclosure_zonotope}).

    Firstly, we derive the gradient $\grad{\nnHiddenSet_{k-1}}{c_{k(i)}}$ needed for $\nnGradSet_{k-1,c}$, for which we need the gradients of center $c_{k(i)} = \approxSlope_{k(i)}\,c_{k-1(i)} + \nicefrac{1}{2}\,(\approxErrorU_{k(i)} + \approxErrorL_{k(i)})$ w.r.t. the input set $\nnHiddenSet_{k-1}$ for each dimension $i\in[\numNeurons_k]$. The image enclosure is applied for each dimension individually; therefore, we can consider each dimension separately because for any dimensions $i,j\in[\numNeurons_{k-1}]$, where $i\neq j$:
    \begin{align}\label{prop:backprop_fast_enclose:eq1}
      \grad{\nnHiddenSet_{k-1(j)}}{c_{k(i)}} & = \shortZ{0}{\zeros}\text{,} & \grad{\nnHiddenSet_{k-1(j)}}{G_{k(i,\cdot)}} & = \shortZ{0}{\zeros}\text{.}
    \end{align}
    Let $i\in[\numNeurons_k]$ be a fixed dimension and $j\in[p]$ a fixed index of a generator.
    We require the gradient of the slope $\approxSlope_{k(i)}$ and the offset $\approxError_{c,k(i)}$. The gradient of the slope $\approxSlope_{k(i)}$ is:
    \begin{align*}
      \partgrad{c_{k-1(i)}}{\approxSlope_{k(i)}} \overset{\text{\cref{def:lin_approx}}}   & {=} \frac{\actfun'(u_{k-1(i)}) - \actfun'(l_{k-1(i)})}{u_{k-1(i)} - l_{k-1(i)}}\text{,}                                                            \\
      \partgrad{G_{k-1(i,j)}}{\approxSlope_{k(i)}} \overset{\text{\cref{def:lin_approx}}} & {=} \left(\frac{\actfun'(u_{k-1(i)}) + \actfun'(l_{k-1(i)}) - 2\,\approxSlope_{k(i)}}{u_{k-1(i)} - l_{k-1(i)}}\right)\,\sign(G_{k-1(i,j)})\text{.}
    \end{align*}

    Let $\xapproxErrorU_k$ and $\xapproxErrorL_k$ be the points of the approximation errors $\approxErrorU_k$ and $\approxErrorL_k$:
    \begin{align*}
      \xapproxErrorU_k & = \argmax_{x\in\extremePoints} \actfun_k(x) - \approxFun_k(x)\text{,} &
      \xapproxErrorL_k & = \argmin_{x\in\extremePoints} \actfun_k(x) - \approxFun_k(x)\text{.}
    \end{align*}
    To prevent repetitions in this proof, let $g$ denote the center, or an arbitrary generator of the input set $\nnHiddenSet_{k-1}$:
    \begin{equation*}
      g\in\left\{c_{k-1},G_{k-1(\cdot,1)},G_{k-1(\cdot,2)},\dots,G_{k-1(\cdot,p)}\right\}\text{.}
    \end{equation*}
    For $(\xapproxError,\approxError)\in\{(\xapproxErrorU_k,\approxErrorU_k),(\xapproxErrorL_k,\approxErrorL_k)\}$, we apply the chain rule and the product-rule to derive the gradients of the approximation error~$\approxError_{(i)}$:
    \begin{align*}
      \partgrad{g_{(i)}}{\approxError_{(i)}} & = \partgrad{g_{(i)}}{\left(\actfun_k(\xapproxError_{(i)}) - \approxSlope_{k(i)}\,\xapproxError_{(i)}\right)} = \left(\actfun'_k(\xapproxError_{(i)}) - \approxSlope_{k(i)}\right)\,\partgrad{g_{(i)}}{\xapproxError_{(i)}} - \partgrad{g_{(i)}}{\approxSlope_{k(i)}}\,\xapproxError_{(i)}\text{.}
    \end{align*}
    Using the gradient of the slope $\approxSlope_{k(i)}$, we derive the gradient of the center $c_{k(i)}$:
    \begin{align*}
      \partgrad{c_{k-1(i)}}{c_{k(i)}}   & = \partgrad{c_{k-1(i)}}{\left(\approxSlope_{k(i)}\,c_{k-1(i)} + \frac{1}{2}\,\left(\approxErrorU_{k(i)} + \approxErrorL_{k(i)}\right)\right)} = \approxSlope_{k(i)} + \partgrad{c_{k-1(i)}}{\approxSlope_{k(i)}}\,c_{k-1(i)} + \frac{1}{2}\,\left(\partgrad{c_{k-1(i)}}{\approxErrorU_{k(i)}} + \partgrad{c_{k-1(i)}}{\approxErrorL_{k(i)}}\right)\text{,} \\
      \partgrad{G_{k-1(i,j)}}{c_{k(i)}} & = \partgrad{G_{k-1(i,j)}}{\left(\approxSlope_{k(i)}\,c_{k-1(i)} + \frac{1}{2}\,\left(\approxErrorU_{k(i)} + \approxErrorL_{k(i)}\right)\right)} = \partgrad{G_{k-1(i,j)}}{\approxSlope_{k(i)}}\,c_{k-1(i)} + \frac{1}{2}\,\left(\partgrad{G_{k-1(i,j)}}{\approxErrorU_{k(i)}} + \partgrad{G_{k-1(i,j)}}{\approxErrorL_{k(i)}}\right)\text{.}
    \end{align*}
    Hence, we have
    \begin{align}\label{eq:prop:bpfastenc:2}
      \begin{split}
        \grad{\nnHiddenSet_{k-1(i)}}{c_{k(i)}} & = \shortZ{\partgrad{c_{k-1(i)}}{c_{k(i)}}}{\partgrad{G_{k-1(i,\cdot)}}{c_{k(i)}}} \\
        & = \approxSlope_{k(i)} + c_{k-1(i)}\,\grad{\nnHiddenSet_{k-1(i)}}{\approxSlope_{k(i)}} + \frac{1}{2}\,\left(\grad{\nnHiddenSet_{k-1(i)}}{\approxErrorU_{k(i)}} + \grad{\nnHiddenSet_{k-1(i)}}{\approxErrorL_{k(i)}}\right)\text{.}
      \end{split}
    \end{align}

    Secondly, we derive the gradient $\grad{\nnHiddenSet_{k-1(i)}}{G_{k(i,j)}}$ needed for $\nnGradSet_{k-1,p}$. Let $j'\in[\numNeurons_k]$ be a different generator index: $j'\neq j$; the gradient of $G_{k(i,j)}$ is
    \begin{align*}
      \partgrad{c_{k-1(i)}}{G_{k(i,j)}}    & = \partgrad{c_{k-1(i)}}{\left(\approxSlope_{k(i)}\,G_{k-1(i,j)}\right)} = \partgrad{c_{k-1(i)}}{\approxSlope_{k(i)}}\,G_{k-1(i,j)}\text{,}                           \\
      \partgrad{G_{k-1(i,j)}}{G_{k(i,j)}}  & = \partgrad{G_{k-1(i,j)}}{\left(\approxSlope_{k(i)}\,G_{k-1(i,j)}\right)} = \approxSlope_{k(i)} + \partgrad{G_{k-1(i,j)}}{\approxSlope_{k(i)}}\,G_{k-1(i,j)}\text{,} \\
      \partgrad{G_{k-1(i,j')}}{G_{k(i,j)}} & = \partgrad{G_{k-1(i,j')}}{\left(\approxSlope_{k(i)}\,G_{k-1(i,j)}\right)} = \partgrad{G_{k-1(i,j')}}{\approxSlope_{k(i)}}\,G_{k-1(i,j)}\text{.}
    \end{align*}
    Hence, we have
    \begin{align}\label{eq:prop:bpfastenc:3}
      \begin{split}
        \grad{\nnHiddenSet_{k-1(i)}}{G_{k(i,j)}} & = \shortZ{\partgrad{c_{k-1(i)}}{G_{k(i,j)}}}{\partgrad{G_{k-1(i,\cdot)}}{G_{k(i,j)}}} = G_{k-1(i,j)}\,\grad{\nnHiddenSet_{k-1(i)}}{\approxSlope_{k(i)}} + \approxSlope_{k(i)}\,\shortZ{\zeros}{\mathbbm{1}_{(i,j)}}\text{,}
      \end{split}
    \end{align}
    where $\mathbbm{1}_{(i,j)}\in\{0,1\}$ is a matrix containing only zeros except for position $(i,j)$ which contains a one.

    Thirdly, we derive $\nnGradSet_{k-1,q}$. Please recall, the diagonal entries of $G_{k(\cdot,p+[\numNeurons_k])}$ contain the approximation errors, while the non-diagonal entries are $0$; hence, the gradient of any non-diagonal entry $j'\in[q]\colon j' > p \wedge j' \neq p + i$ is $0$: $\grad{\nnHiddenSet_{k-1(i)}}{G_{k(i,j')}} = \shortZ{0}{\zeros}$.
    Hence, we can simplify $\nnGradSet_{k-1,q}$:
    \begin{align}\label{eq:prop:bpfastenc:5}
      \begin{split}
        \nnGradSet_{k-1,q(i)} & = \sum_{j=p+1}^{q} G'_{k(i,j)}\,\grad{\nnHiddenSet_{k-1(i)}}{G_{k(i,j)}} = G'_{k(i,p + i)}\,\frac{1}{2}\,\left(\grad{\nnHiddenSet_{k-1(i)}}{\approxErrorU_{k(i)}} - \grad{\nnHiddenSet_{k-1(i)}}{\approxErrorL_{k(i)}}\right)\text{.}
      \end{split}
    \end{align}
    We add $\nnGradSet_{k-1,c}$ and $\nnGradSet_{k-1,p}$ followed by reordering the terms:
    \begin{align}\label{eq:prop:bpfastenc:4}
      \begin{split}
        \nnGradSet_{k-1,c(i)} + \nnGradSet_{k-1,p(i)} & = c'_{k(i)}\,\grad{\nnHiddenSet_{k-1(i)}}{c_{k(i)}} + \sum_{j=1}^{p} G'_{k(i,j)}\,\grad{\nnHiddenSet_{k-1(i)}}{G_{k(i,j)}}                                                         \\
        & \overset{\text{\cref{eq:prop:bpfastenc:2}}}{=} c'_{k(i)}\,\left(\approxSlope_{k(i)} + c_{k-1(i)}\,\grad{\nnHiddenSet_{k-1(i)}}{\approxSlope_{k(i)}} + \frac{1}{2}\,\left(\grad{\nnHiddenSet_{k-1(i)}}{\approxErrorU_{k(i)}} + \grad{\nnHiddenSet_{k-1(i)}}{\approxErrorL_{k(i)}}\right)\right)          \\
        & \quad + \sum_{j=1}^{p} G'_{k(i,j)}\,\grad{\nnHiddenSet_{k-1(i)}}{G_{k(i,j)}}   \\
        & \overset{\text{\cref{eq:prop:bpfastenc:3}}}{=} c'_{k(i)}\,\left(\approxSlope_{k(i)} + c_{k-1(i)}\,\grad{\nnHiddenSet_{k-1(i)}}{\approxSlope_{k(i)}} + \frac{1}{2}\,\left(\grad{\nnHiddenSet_{k-1(i)}}{\approxErrorU_{k(i)}} + \grad{\nnHiddenSet_{k-1(i)}}{\approxErrorL_{k(i)}}\right)\right)          \\
        & \quad + \sum_{j=1}^{p} G'_{k(i,j)}\,\left(G_{k-1(i,j)}\,\grad{\nnHiddenSet_{k-1(i)}}{\approxSlope_{k(i)}} + \approxSlope_{k(i)}\,\shortZ{\zeros}{\mathbbm{1}_{(i,j)}}\right)   \\
        & = \approxSlope_{k(i)}\,\nnGradSet_{k(i)} + \left(c_{k-1(i)}\,c'_{k(i)} + G^\top_{k-1(i,\cdot)}\,G'_{k(i,[p])}\right)\,\grad{\nnHiddenSet_{k-1(i)}}{\approxSlope_{k(i)}} \\
        & \quad + c'_{k(i)}\,\frac{1}{2}\,\left(\grad{\nnHiddenSet_{k-1(i)}}{\approxErrorU_{k(i)}} + \grad{\nnHiddenSet_{k-1(i)}}{\approxErrorL_{k(i)}}\right)\text{.}
      \end{split}
    \end{align}
    Finally, we obtain
    \begin{align*}
                                                  & \nnGradSet_{k-1(i)} \overset{\text{\cref{eq:prop:bpfastenc:1}}} {=} \nnGradSet_{k-1,c(i)} + \nnGradSet_{k-1,p(i)} + \nnGradSet_{k-1,q(i)}                                                                                                                                                                 \\
      \overset{\text{\cref{eq:prop:bpfastenc:4}}} & {=} \approxSlope_{k(i)}\,\nnGradSet_{k(i)} + \left(c_{k-1(i)}\,c'_{k(i)} + G'_{k(i,[p])}\,G_{k-1(i,\cdot)}^\top\right)\,\grad{\nnHiddenSet_{k-1(i)}}{\approxSlope_{k(i)}} + c'_{k(i)}\,\grad{\nnHiddenSet_{k-1(i)}}{\approxError_{c,k(i)}} + \nnGradSet_{k-1,q(i)}                                        \\
      \overset{\text{\cref{eq:prop:bpfastenc:5}}} & {=} \approxSlope_{k(i)}\,\nnGradSet_{k(i)} + \left(c_{k-1(i)}\,c'_{k(i)} + G'_{k(i,[p])}\,G_{k-1(i,\cdot)}^\top\right)\,\grad{\nnHiddenSet_{k-1(i)}}{\approxSlope_{k(i)}}                                                                                                                                 \\
                                                  & \quad + c'_{k(i)}\,\frac{1}{2}\,\left(\grad{\nnHiddenSet_{k-1(i)}}{\approxErrorU_{k(i)}} + \grad{\nnHiddenSet_{k-1(i)}}{\approxErrorL_{k(i)}}\right) + G'_{k(i,p + i)}\,\frac{1}{2}\,\left(\grad{\nnHiddenSet_{k-1(i)}}{\approxErrorU_{k(i)}} - \grad{\nnHiddenSet_{k-1(i)}}{\approxErrorL_{k(i)}}\right) \\
                                                  & = \approxSlope_{k(i)}\,\nnGradSet_{k(i)} + \left(c_{k-1(i)}\,c'_{k(i)} + G'_{k(i,[p])}\,G_{k-1(i,\cdot)}^\top\right)\,\grad{\nnHiddenSet_{k-1(i)}}{\approxSlope_{k(i)}}                                                                                                                                   \\
                                                  & \quad + \frac{1}{2}\,\left(c'_{k(i)} + G'_{k(i,p + i)}\right)\,\grad{\nnHiddenSet_{k-1(i)}}{\approxErrorU_{k(i)}} + \frac{1}{2}\,\left(c'_{k(i)} - G'_{k(i,p + i)}\right)\,\grad{\nnHiddenSet_{k-1(i)}}{\approxErrorL_{k(i)}}\text{.}\qedhere
    \end{align*}
  \end{proof}
\end{repproposition}

\begin{repproposition}{prop:set_backprop}[Set-Based Backpropagation]
  \propSetBackprop
  \begin{proof}
    If $k=\numLayers$, we compute the gradient of the set-based loss according to \cref{prop:grad_set_loss}. We assume $k<\numLayers$.
    Let $\nnGradSet_{k} = \shortZ{c'_{k}}{G'_{k}}$ and $\nnHiddenSet_{k} = \shortZ{c_{k}}{G_{k}}$.

    We split cases on the type of the $k$-th layer and simplify the terms.
    \begin{enumerate}[label=Case (\roman*)., wide=0pt, font=\itshape]
      \item The $k$-th layer is linear. For dimension $i\in[\numNeurons_k]$, we have
            \begin{align}\label{eq:prop:set_backprop:1}
              \begin{split}
                \grad{\nnHiddenSet_{k-1}}{c_{k(i)}}   & = \grad{\nnHiddenSet_{k-1}}{\left(W_{k(i,\cdot)}\,c_{k-1} + b_{k(i)}\right)} = \shortZ{W_{k(i,\cdot)}^\top}{\zeros}\text{,}                         \\
                \grad{\nnHiddenSet_{k-1}}{G_{k(i,j)}} & = \grad{\nnHiddenSet_{k-1}}{\left(W_{k(i,\cdot)}\,G_{k-1(\cdot,j)}\right)} = \shortZ{\zeros}{W_{k(i,\cdot)}^\top\,e_j^\top}\text{.}
              \end{split}
            \end{align}
            Thus,
            \begin{align*}
              \nnGradSet_{k-1} \overset{\text{\cref{eq:grad-chain-rule-part-der}}} & {=} \sum_{i=1}^{\numNeurons_k} c'_{k(i)}\,\grad{\nnHiddenSet_{k-1}}{c_{k(i)}} + \sum_{i=1}^{\numNeurons_k}\sum_{j=1}^{q}  G'_{k(i,j)}\,\grad{\nnHiddenSet_{k-1}}{G_{k(i,j)}}             \\
              \overset{\text{\cref{eq:prop:set_backprop:1}}}                       & {=} \sum_{i=1}^{\numNeurons_k} c'_{k(i)}\,\shortZ{W_{k(i,\cdot)}^\top}{\zeros} + \sum_{i=1}^{\numNeurons_k} \sum_{j=1}^{q} G'_{k(i,j)}\,\shortZ{\zeros}{W_{k(i,\cdot)}^\top\,\,e_j^\top} \\
                                                                                   & = \shortZ{\sum_{i=1}^{\numNeurons_k} W_{k(i,\cdot)}^\top\,c'_{k(i)}}{\sum_{i=1}^{\numNeurons_k} W_{k(i,\cdot)}^\top\,G'_{k(i,\cdot)}}                                                    \\
                                                                                   & = \shortZ{W_k^\top\,c'_{k}}{W_k^\top\,G'_{k}}                                                                                                                                            \\
                                                                                   & = W_k^\top\,\nnGradSet_k\text{.}
            \end{align*}
      \item The $k$-th layer is nonlinear. \cref{prop:backprop_fast_enclose} proves the correctness.
    \end{enumerate}
  \end{proof}
\end{repproposition}

\begin{repproposition}{prop:set-weight-update}[Gradient Set w.r.t. Weights and Bias]
  \propSetWeightUpdate
  \begin{proof}
    We rewrite the gradient by applying the chain rule for partial derivatives:
    \begin{align*}
      \grad{W_{k}}{\setLoss(\nnTarget,\nnOutputSet)} & = \sum_{i=1}^{\numNeurons_k} c'_{k(i)}\,\grad{W_{k}}{c_{k(i)}} + \sum_{i=1}^{\numNeurons_k} \sum_{j=1}^{q} G'_{k(i,j)}\,\grad{W_{k}}{G_{k(i,j)}}\text{,} \\
      \grad{b_{k}}{\setLoss(\nnTarget,\nnOutputSet)} & = \sum_{i=1}^{\numNeurons_k} c'_{k(i)}\,\grad{b_{k}}{c_{k(i)}} + \sum_{i=1}^{\numNeurons_k} \sum_{j=1}^{q} G'_{k(i,j)}\,\grad{b_{k}}{G_{k(i,j)}}\text{,}
    \end{align*}
    where $G_{k}\in\R^{\numNeurons_k\times q}$. Moreover, we have for dimension $i\in[\numNeurons_k]$ and generator index $j\in[q]$:
    \begin{align*}
      \grad{W_{k}}{c_{k(i)}}   & = \grad{W_{k}}{\left(W_{k(i,\cdot)}\,c_{k-1} + b_{k(i)}\right)} = e_i\,c_{k-1}^\top\text{,}        &
      \grad{W_{k}}{G_{k(i,j)}} & = \grad{W_{k}}{\left(W_{k(i,\cdot)}\,G_{k-1(\cdot,j)}\right)} = e_i\,G_{k-1(\cdot,j)}^\top\text{,}   \\
      \grad{b_{k}}{c_{k(i)}}   & = \grad{b_{k}}{\left(W_{k(i,\cdot)}\,c_{k-1} + b_{k(i)}\right)} = e_i\text{,}                      &
      \grad{b_{k}}{G_{k(i,j)}} & = \grad{b_{k}}{\left(W_{k(i,\cdot)}\,G_{k-1(\cdot,j)}\right)} = \zeros\text{,}
    \end{align*}
    where $e_i\in\{0,1\}^{\numNeurons_k}$ is the $i$-th standard basis vector. Thus,
    \begin{align*}
      \grad{W_{k}}{\setLoss(\nnTarget,\nnOutputSet)} & = c'_{k}\,c_{k-1}^\top + G'_{k}\,G_{k-1}^\top = \nnGradSet_k\odot\nnHiddenSet_{k-1}^\top\text{,} &
      \grad{b_{k}}{\setLoss(\nnTarget,\nnOutputSet)} & = c'_{k} = \nnGradSet_k\odot\shortZ{1}{\zeros}^\top\text{.}\qedhere
    \end{align*}
  \end{proof}
\end{repproposition}

\begin{repproposition}{prop:fast_enclose_time_complexity}
  \propFastEncloseTimeComplexity
  \begin{proof}
    Finding the interval bounds of $\nnHiddenSet_{k-1}$ (\cref{line:alg_fast_img_enc:bounds}) takes time $\bigO(n\,q)$ (\cref{prop:interval_enclosure_zonotope}). Computing the linear approximation (\cref{line:alg_fast_img_enc:lin_approx}) and the approximation errors (\cref{line:alg_fast_img_enc:approx_errors}) for each neuron takes constant time; hence, the loop takes time $\bigO(n)$. The linear map of $\nnHiddenSet_{k-1}$ (\cref{line:alg_fast_img_enc:lin_map}) takes time $\bigO(n^2\,q)$ (\cref{prop:linear_map_zonotope}). Adding the approximation errors (\cref{line:alg_fast_img_enc:mink}) takes time $\bigO(n)$. Thus, in total we have $\bigO(n\,q) + \bigO(n) + \bigO(n^2\,q) + \bigO(n) = \bigO(n^2\,q)$.
  \end{proof}
\end{repproposition}

\changed{
  \begin{repproposition}{prop:set_forward_complexity}
    \propSetForwardPropComplexity
    \begin{proof}
      The initial $\epsilon$-perturbance set has $\numNeurons_0$ generators (\cref{line:alg_set_train:init_perturb_input}) and every nonlinear layer adds $\numNeurons_k$ new generators for the approximation errors (\cref{algo:fast_enclose}). Moreover, there are at most $\numLayers$ nonlinear layers. Thus, in total, there are at most $(\numNeurons_0 + \sum_{k\in[\numLayers]}\numNeurons_k)$ generators.

      Time Complexity: The $k$-th step of the set-based forward propagation takes time $\bigO(\numNeurons_\text{max}^2\,q)$: The linear map (\cref{line:alg_set_train:forward_lin}) as well as the image enclosure (\cref{line:alg_set_train:forward_act}) takes time $\bigO(\numNeurons_{k-1}\,\numNeurons_k\,q) = \bigO(\numNeurons_\text{max}^2\,q)$ (\cref{prop:linear_map_zonotope,prop:fast_enclose_time_complexity}). Thus, the set-based forward propagation~(\crefrange{line:alg_set_train:forward_prop_start}{line:alg_set_train:forward_prop_end}) takes time $\numLayers\,\bigO(\numNeurons_\text{max}^2\,q)$.
    \end{proof}
  \end{repproposition}
}

\begin{repproposition}{prop:set_training_complexity}
  \propSetTrainingComplexity
  \begin{proof}
    The set-based forward propagation~(\crefrange{line:alg_set_train:forward_prop_start}{line:alg_set_train:forward_prop_end}) takes time $\numLayers\,\bigO(\numNeurons_\text{max}^2\,q)$~(\cref{prop:set_forward_complexity}).
    The gradient of the set-based loss has $(\numNeurons_\numLayers + \numNeurons_\numLayers\,q)$ entries, and the computation of each entry takes constant time; hence, computing the gradient takes time $\bigO(\numNeurons_\numLayers + \numNeurons_\numLayers\,q)$.
    The $k$-th step of the set-based backpropagation takes at most $\bigO(\numNeurons_{k-1}\,\numNeurons_k\,q) = \bigO(\numNeurons_\text{max}^2\,q)$ time: a linear layer computes a linear map (\cref{line:alg_set_train:backward_lin}), which takes time $\bigO(\numNeurons_{k-1}\,\numNeurons_k\,q)$~(\cref{prop:linear_map_zonotope}), and the set-based backpropagation of an image enclosure (\cref{line:alg_set_train:backward_act}) takes time $\bigO(\numNeurons_{k-1}\,\numNeurons_k\,q)$~(\cref{prop:backprop_fast_enclose}). Hence, the set-based backpropagation (\crefrange{line:alg_set_train:back_prop_start}{line:alg_set_train:back_prop_end}) takes time $\numLayers\,\bigO(\numNeurons_\text{max}^2\,q)$.
    Updating a weight matrix takes time $\bigO(\numNeurons_{k-1}\,\numNeurons_k + \numNeurons_{k-1}\,\numNeurons_k\,q) = \bigO(\numNeurons_\text{max}^2\,q)$ (\cref{line:alg_set_train:backward_weight_update}) and updating a bias vector takes time $\bigO(\numNeurons_k) = \bigO(\numNeurons_\text{max})$ (\cref{line:alg_set_train:backward_bias_update}). There are at most $\numLayers$ linear layers; hence, updating the weight matrix and bias vector of all linear layers takes time $\numLayers\,(\bigO(\numNeurons_\text{max}^2\,q) + \bigO(\numNeurons_\text{max})) = \numLayers\,\bigO(\numNeurons_\text{max}^2\,q)$.
    Thus, in total, an iteration of set-based training takes time $3\,\numLayers\,\bigO(\numNeurons_\text{max}^2\,q\,\numLayers) = \bigO(\numNeurons_\text{max}^2\,q\,\numLayers)$.
  \end{proof}
\end{repproposition}

\end{document}